\documentclass[11pt,a4paper]{article}

\usepackage[utf8]{inputenc}
\usepackage[T1]{fontenc}
\usepackage{amsmath,amssymb,amsthm}
\usepackage{mathtools}
\usepackage{bm}
\usepackage{graphicx}
\usepackage{booktabs}
\usepackage{algorithm}
\usepackage{algorithmic}
\usepackage{hyperref}
\usepackage{cleveref}
\usepackage{natbib}
\usepackage{geometry}
\usepackage{enumitem}
\usepackage{xcolor}
\usepackage{tikz}
\usetikzlibrary{arrows.meta,positioning,shapes.geometric}

\newcommand{\R}{\mathbb{R}}
\newcommand{\N}{\mathbb{N}}
\newcommand{\E}{\mathbb{E}}

\newcommand{\bx}{\boldsymbol{x}}
\newcommand{\bc}{\boldsymbol{c}}
\newcommand{\ba}{\mathbf{a}}
\newcommand{\bz}{\boldsymbol{z}}

\newcommand{\bu}{\boldsymbol{u}}
\newcommand{\be}{\boldsymbol{e}}
\newcommand{\bS}{\boldsymbol{S}}
\newcommand{\bE}{\mathbf{E}}
\newcommand{\bW}{\boldsymbol{W}}

\newcommand{\bbeta}{\boldsymbol{\beta}}
\newcommand{\bs}{\boldsymbol{s}}
\newcommand{\bb}{\boldsymbol{b}}
\newcommand{\bA}{\boldsymbol{A}}
\newcommand{\bB}{\boldsymbol{B}}
\newcommand{\bI}{\boldsymbol{I}}

\DeclareMathOperator{\LayerNorm}{LayerNorm}
\DeclareMathOperator{\Flatten}{Flatten}

\title{%
\textbf{Tab-TRM: Tiny Recursive Model for\\[0.5em]
Insurance Pricing on Tabular Data}
}

\author{%
Kishan Padayachy\footnote{insureAI, kishan@insureai.co} \and
Ronald Richman\footnote{insureAI, ronaldrichman@gmail.com}
\and Mario V.~W\"uthrich\footnote{Department of Mathematics, ETH Zurich,
mario.wuethrich@math.ethz.ch}
}

\date{\today}

\begin{document}

\maketitle

\begin{abstract}
\noindent
We introduce Tab-TRM (Tabular-Tiny Recursive Model), a network architecture that adapts the recursive latent reasoning paradigm of Tiny Recursive Models (TRMs) to insurance modeling. Drawing inspiration from both the Hierarchical Reasoning Model (HRM) and its simplified successor TRM, the Tab-TRM model makes predictions by reasoning over the input features. It maintains two learnable latent tokens - an answer token and a reasoning state - that are iteratively refined by a compact, parameter-efficient recursive network. The recursive processing layer repeatedly updates the reasoning state given the full token sequence and then refines the answer token, in close analogy with iterative insurance pricing schemes.  Conceptually, Tab-TRM bridges classical actuarial workflows - iterative generalized linear model fitting and minimum-bias calibration - on the one hand, and modern machine learning, in terms of Gradient Boosting Machines, on the other.

\medskip

\noindent
\textbf{Keywords:} Tiny Recursive Models, Hierarchical Reasoning, Insurance Pricing, Regression, Tabular Deep Learning, Credibility Theory, Recursive Neural Networks
\end{abstract}

\section{Introduction}
\label{sec:intro}
Many actuarial prediction problems such as pricing, reserving and experience analysis in insurance are closely related to the field of tabular machine learning; this connection has been increasingly explored in recent years. Over roughly the past decade, a growing actuarial literature has explored deep learning as a complement or alternative to classical generalized linear models (GLMs); see, for example, \citet{wuethrichmerz2023statistical} for an overview of actuarial learning tools, \citet{richman2021ai1,richman2021ai2} for a two-part review of AI in actuarial science, and \citet{wuethrich2025aitools} for a wide ranging actuarial treatment of these tools.

A foundational contribution in tabular machine learning was made by \citet{guo2016entity}, who showed that categorical variables can be mapped into low-dimensional continuous \emph{entity embeddings} learned jointly with the prediction task. This approach addresses the high dimensionality and sparsity of one-hot encodings of categorical variables, it allows similar categories to cluster in the embedding space, and it improves generalization - properties that are particularly valuable in insurance where high-cardinality categorical features such as vehicle type, occupation and region are ubiquitous. Building on this idea, early actuarial work demonstrated that feed-forward neural networks (FNNs) with entity embeddings for categorical covariates can be viewed as GLMs on top of learned feature transformations \citep{richman2021ai2}. With the right exposure handling, log-links and deviance-based loss functions, such architectures already deliver strong performance on pricing benchmarks while preserving core actuarial structures such as multiplicative tariffs and offsets.

In parallel, gradient boosting machines (GBMs) - in particular XGBoost - have become a de facto benchmark for predictive performance on tabular insurance data. \citet{noll2020case} compared GLMs, FNNs and GBMs on a French motor third-party liability (MTPL) portfolio and found that GBMs and neural networks consistently outperform GLMs in terms of Poisson deviance scoring, while remaining computationally tractable. GBMs excel at capturing complex feature interactions through sequential tree construction, though they lack the smooth gradient-based optimization and representation-learning capabilities of FNNs.

Several architectures aim to combine the interpretability of GLMs with the flexibility of neural networks. The \emph{Combined Actuarial Neural Network} (CANN) of \citet{schelldorfer2019nesting} nests a classical GLM within a neural network via a skip-connection, allowing the network to learn residual corrections to the GLM baseline. This structure ensures that the model never performs worse than the GLM and can systematically identify missing interactions. The \emph{LocalGLMnet} of \citet{richman2023localglmnet} takes a different approach: it allows the GLM regression coefficients themselves to be feature-dependent functions learned by a neural network. This yields a model that behaves locally as a GLM - with interpretable, policy-specific coefficients - while globally capturing nonlinear effects and interactions.

More recently, several architectures have explicitly imported modern representation-learning ideas into actuarial modeling while preserving key actuarial concepts. The \emph{Credibility Transformer} adapts the Transformer architecture to tabular insurance data by embedding each covariate as a token and introducing a special \emph{credibility token}. This token combines global portfolio information with observation-specific information through a credibility-weighted average, mirroring B\"uhlmann credibility inside an attention mechanism \citep{richman2025credibilitytransformer}. 
Building on the Credibility Transformer, the In-Context Learning Enhanced Credibility Transformer (ICL-CT) attaches a \emph{context batch} of similar insurance risks to each target policy. An in-context learning mechanism then allows the model to adapt its internal representation to local risk patterns and even generalize to unseen categorical levels \citep{padayachy2025iclcredibility}; in fact, this idea is very close to the foundation models used in large language models (LLMs). These examples illustrate a general pattern: deep learning models tailored to actuarial structure - exposure offsets, credibility, tariff factors - can be both accurate and interpretable.

In parallel with the actuarial deep-learning literature, LLMs have driven a rapid resurgence of interest in machine reasoning. A key observation is that LLMs which struggle with multi-step computations can dramatically improve when trained to emit intermediate steps into a ``scratchpad'' before producing the final answer \citep{nye2021scratchpads}. This idea, popularized as \emph{chain-of-thought} (CoT) prompting \citep{wei2022cot}, has inspired a range of post-training methods and test-time compute strategies for enhancing reasoning; see Appendix~\ref{app:llm_background} for details.

A complementary direction, pursued by the Hierarchical Reasoning Model (HRM) and Tiny Recursive Model (TRM), is to perform iterative refinement in a \emph{latent} space rather than through explicit CoT text.\footnote{An interesting parallel to this is credibility; usually credibility is applied in the output space of rates, but in \cite{richman2025credibilitytransformer} it is applied in the latent embedding space of the model.} HRM couples two Transformer-based recurrent modules that operate at different effective time scales: a fast low-level module and a slower high-level module. By recursing over a small number of latent tensors, using deep supervision across multiple ``supervision steps'' and an adaptive computation time mechanism, HRM achieves large effective depth with modest parameter count and demonstrates strong performance on Sudoku-Extreme, Maze-Hard and ARC-AGI, using only around 1,000 training examples and no pre-training or CoT supervision \citep{wang2025hrm}. The \emph{Tiny Recursive Model} (TRM) of \citet{jolicoeur2025trm} provides a compelling simplification: it replaces the two-level structure with a single tiny network that jointly updates an embedded candidate answer $\ba$ and an auxiliary latent state $\bz$ that acts as a scratchpad of intermediate computations. At each reasoning step, the model first refines $\bz$ given the input $\bx$, the current answer $\ba$, and the current state $\bz$, and then proposes an improved answer $\ba$ conditioned on the updated $\bz$. TRM explicitly avoids fixed-point gradient approximations and the extra forward passes required by adaptive computation time in HRM, yet substantially improves on HRM across these benchmarks.

\subsection{Motivation and Research Gap}

From the perspective of actuarial science, LLM-based reasoning is conceptually important - it shows that explicit, multi-step computation can be learned and exploited by neural network models - but the approaches are not directly applicable as a drop-in solution for insurance pricing. Actuaries typically seek compact architectures that can be trained on a single portfolio without access to web-scale CoT data, reinforcement learning infrastructure, or the computational budgets required by test-time compute (TTC). At the same time, the grid-based benchmarks on which HRM and TRM excel share interesting structural parallels with actuarial tabular data: (i)~inputs are naturally decomposed into a small, fixed set of tokens (grid cells or tariff factors); (ii)~the prediction task requires combining evidence from multiple tokens to produce a single output; and (iii)~iterative refinement - whether of Sudoku cell values or tariff relativities - is a natural inductive bias, well exploited by GBMs.

To date, both HRM and TRM have been evaluated almost exclusively on structured grid-based reasoning problems, where inputs and outputs are discrete symbols and the objective is exact sequence-to-sequence prediction. Here, we explore whether the same ideas - latent recursion over a small set of tokens, deep supervision, and tiny networks - translate to traditional tabular modeling problems, where the goals are accurate point forecasts for insurance pricing. We utilize the fact that tabular data have several structural features that are particularly well-suited to the TRM philosophy: (i)~each feature can naturally be treated as a separate token; (ii)~the number of tokens is small and fixed for a given insurance tariff structure; and (iii)~a model that repeatedly refines a low-dimensional answer representation $\ba$ and a global latent scratchpad $\bz$ provides a natural inductive bias for capturing complex feature interactions without resorting to very wide or very deep FNNs.

\subsection{Contributions}

In this work we bring TRM-style latent recursion to actuarial tabular modeling. Our contributions are fourfold. First, we present, to our knowledge, the first adaptation of Tiny Recursive Models to insurance pricing or tabular machine learning. In the context of pricing, we represent each policy as a short sequence of feature tokens augmented with answer and latent reasoning tokens and introduce a practical architecture for TRM-style tabular modeling with a ``tiny'' recursive processing layer that repeatedly updates a global latent state and an answer token. Third, we provide empirical validation on the classical benchmark of the  French MTPL data of \cite{Dutang}, and we show that Tab-TRM achieves a strong Poisson deviance score with a very small network. Finally, we relate the advantages of tiny recursive reasoning models - previously demonstrated on synthetic and symbolic puzzles - to noisy, mixed-type tabular data, and we discuss how the architecture sits alongside other architectures such as the credibility-based Transformers and in-context learning architectures, highlighting parallels with iterative GLM fitting and GBMs.

To make Tab-TRM accessible to actuarial practitioners, it is helpful to emphasise parallels with familiar workflows. The recursive refinement of $(\ba,\bz)$ can be viewed as a \emph{learned analogue of iterative GLM fitting}. Instead of Iteratively Reweighted Least Squares (IRLS), a tiny shared network learns how to map a ``working predictor'' and latent tariff structure to a slightly improved one. Each recursion step is analogous to another pass through the portfolio, updating relativities and rebalancing as indications and constraints are reassessed. Another analogy is to Recurrent Neural Networks (RNNs): if one ``unrolls'' the recursive core over its inner and outer iterations, the effective computation closely resembles that of a RNN unrolled in time, but with two key differences: the state space is explicitly split into an answer component and a reasoning component, and deep supervision can in principle be applied at several checkpoints rather than relying on very long backpropagation through time. In this sense, Tab-TRM can be seen as a bridge between classical actuarial algorithms and RNNs that iterate hidden states over time.

\paragraph{Structure of the paper.}
The remainder of this paper is organised as follows. Section~\ref{sec:methodology} presents the Tab-TRM architecture, covering the tokenization of tabular inputs, the recursive reasoning core, and the Poisson decoder. Section~\ref{sec:interpretation} provides complementary interpretations of Tab-TRM, including connections to state-space models and gradient boosting. Section~\ref{sec:example} demonstrates the methodology on the benchmark French MTPL portfolio, detailing feature encodings and hyper-parameter optimization. Section~\ref{sec:results} reports empirical results, including Poisson deviance scores, interpretability diagnostics, and analysis of the learned representations. Section~\ref{sec:linear_tabtrm} explores a fully linearized variant of Tab-TRM and provides a state-space analysis of the learned dynamics. Section~\ref{sec:discussion} relates Tab-TRM to classical actuarial workflows and other network architectures. Section~\ref{sec:conclusion} concludes and outlines directions for future work. Supplementary material in the appendix includes algorithm pseudo code and hyper-parameter search ranges.

\section{Methodology}
\label{sec:methodology}

In this section we formalize the non-life insurance claims frequency prediction problem, describe how the TRMs  of \citet{jolicoeur2025trm} are adapted to tabular insurance data, and present the resulting Tab-TRM architecture and its training procedure.

\subsection{Problem Formulation - Poisson Claims Frequencies}
To keep things simple we focus on non-life insurance claims frequency modeling using the Poisson model, and the ideas presented here can easily be adapted to other actuarial forecast problems.
The goal is to estimate the expected claim frequency for each insurance policy, conditional on observed risk characteristics (covariates, features). Formally, for each insurance policy $i = 1,\dots,n$, we observe the tuple
\[
(\bx_i, \bc_i, v_i, y_i),
\]
where $\bx_i \in \R^{p_r}$ is a vector of continuous covariates, $\bc_i \in \N^{p_c}$ is a vector of categorical covariates, $v_i > 0$ is the exposure (time at risk), and $y_i \in \N_0$ is the observed claim count (response). Our objective is to learn a regression function $F_\theta \colon \R^{p_r} \times \N^{p_c} \to \R_{>0}$ that maps from the covariate space to the
positive real line, such that $F_\theta(\bx_i, \bc_i)$ gives us a good approximation to the true (unknown) expected claim frequency $\lambda(\bx_i, \bc_i)$ of policyholder $i$, that is,
\begin{equation}
\E[Y_i \mid \bx_i, \bc_i, v_i] = v_i \, \lambda(\bx_i, \bc_i),
\qquad \text{ for all $i=1,\ldots, n$.}
\label{eq:conditional_expectation}
\end{equation}
Lifting this to a full statistical model, we furthermore assume a Poisson model for the claim counts
\begin{equation}
Y_i \mid \bx_i, \bc_i, v_i  \sim \mathrm{Poisson}\left( v_i \,
\lambda(\bx_i, \bc_i)\right).
\label{eq:poisson_model}
\end{equation}
We then try to approximate $\lambda(\cdot)$ from a parametrized class of regression functions $\{F_\theta(\cdot)\}_\theta$.
For example, in the classical GLM setting with log-link function, we have the parametrized
class $\log (F_\theta(\bx_i, \bc_i)) = \beta_0 + \bx_i^\top \bbeta_r + \sum_j \gamma_{c_{ij}}$, with intercept $\beta_0$, $\bbeta_r$ are the regression coefficients for the continuous covariates and $\gamma_{c_{ij}}$ are factor-level effects for the categorical covariates (typically implemented by dummy coding), and $\theta$ collects all these model parameters. Here, we instead model and learn $F_\theta(\cdot)$ via a TRM architecture while retaining the Poisson likelihood structure~\eqref{eq:poisson_model} with the exposure offset. The TRM outputs estimates $F_{\widehat{\theta}}(\bx_i, \bc_i)$, and the model is trained by minimizing the Poisson deviance loss, which corresponds to maximum likelihood estimation under the Poisson assumption \eqref{eq:poisson_model}, assuming independence between the
policies $i=1,\ldots, n$.

\subsection{From HRM to TRM to Tab-TRM}

The HRM of \citet{wang2025hrm} is a recurrent Transformer architecture with two interacting latent states $(\bz_L,\bz_H)$, updated at different effective time scales by two networks $f_L$ and $f_H$. Through repeated application of these networks, together with a one-step gradient approximation derived from deep equilibrium models, a HRM achieves very large effective depth with modest parameter count and shows strong performance on algorithmic benchmarks; we also refer to \citet{bai2019deq} for the underlying equilibrium-gradient ideas. Two additional ingredients are important for HRMs. First, \emph{deep supervision} is used: the model is trained over multiple ``supervision steps'', with reuse of the latent states $(\bz_L,\bz_H)$ across steps and per-step losses, which effectively emulate a very deep network without long backpropagation-through-time. Second, an \emph{adaptive computational time (ACT)} mechanism is introduced, whereby a Q-learning based halting head decides how many supervision steps to apply to each example, with an extra forward pass used to define the continuation target.

TRMs of  \citet{jolicoeur2025trm} show that much of HRM's benefit can be retained, and in fact improved,  with a much simpler design. TRMs reinterpret the two HRM latent states as an embedded solution or \emph{answer token} $\ba$ and an auxiliary \emph{latent reasoning token} $\bz$. A single tiny network is then applied recursively in two roles: repeatedly refining $\bz$, given $(\bx,\ba,\bz)$, and then updating $\ba$, given $(\ba,\bz)$. The recursion is organised into an inner loop of $m$ {\it latent reasoning updates} of $\bz$, an outer loop of $T$ {\it answer updates} of $\ba$, and a set of $N_{\text{sup}}$ repeated passes with deep supervision, in which $(\ba,\bz)$ are carried across passes and detached from the computational graph. TRM abandons the HRM fixed-point gradient approximation and instead backpropagates through a full recursion block. This change substantially improves generalization on the same puzzle benchmarks while still using a single two-layer network with only a few million parameters.

We now introduce Tab-TRM, a specialization of the TRM framework to regression problems on tabular insurance data. Let $\bx \in \R^{p_r}$ and $\bc \in \N^{p_c}$ denote the continuous and categorical covariates for a single policy. Tab-TRM constructs the regression function $F_\theta(\bx, \bc)$ by maintaining two latent vectors per policy: an \emph{answer embedding} $\ba \in \R^{d_a}$ and a \emph{reasoning embedding} $\bz \in \R^{d_z}$, both initialized as learned parameters and prepended as prefix tokens to the input sequence. Given an input representation formed by embedding each covariate component into a $d$-dimensional space, the full sequence takes the form 
\begin{equation*}
[\ba, \bz, \be_1, \ldots, \be_L] ~\in~ \R^{d_a} \times \R^{d_z} \times \R^{d \times L},
\end{equation*}
where $L = p_r + p_c$, and $\be_j \in \R^d$ represents the $d$-dimensional embedding of the $j$-th covariate component, for details on embeddings we also refer to 
\citet{richman2025credibilitytransformer}. 

The Tab-TRM applies a recursive update scheme comprising $T$ outer iterations of answer refinements, each containing $m$ inner iterations of reasoning refinements.
Formally, let $f_z$ and $f_a$ denote the reasoning and answer update networks, respectively. We flatten the current sequence to a vector and apply $f_z$ to update the reasoning token, then use the refined reasoning token alongside the answer token as input to $f_a$. At outer step $t\ge 0$ and inner step $0\le s \le m-1$, the updates are
\begin{align}
\bz^{(t,s+1)} &= \bz^{(t,s)} + f_z\left(\operatorname{Flatten}\left([\ba^{(t)}, \bz^{(t,s)}, \be_1, \ldots, \be_L]\right)\right), \label{eq:z_update} \\
\ba^{(t+1)} &= \ba^{(t)} + f_a\left(\operatorname{Concat}\left(\ba^{(t)}, \bz^{(t,m)}\right)\right), \label{eq:y_update}
\end{align}
where $\bz^{(t,m)}$ is the reasoning state after all $m$ inner iterations, and we initialize the inner loop by $\bz^{(t,0)}=\bz^{(t-1,m)}$. Both updates are residual connections. The final answer embedding $\ba^{(T)}$ is passed through a linear head followed by an exponential activation to produce estimate $F_\theta(\bx, \bc) \in \R_{>0}$, preserving the log-link structure of Poisson GLMs.

Finally, the TRM output $F_\theta(\bx, \bc)$ is multiplied with 
 the exposure $v_i$, and the model is trained by minimizing the Poisson deviance loss. Three architectural departures from the original TRM are noteworthy. First, we employ separate networks $f_z$ and $f_a$ rather than a single shared network, since in the tabular setting the $\bz$-update conditions on all tokens (including the covariates), whereas the $\ba$-update conditions only on the current answer and reasoning states $(\ba,\bz)$. Second, we replace the attention-based message passing of TRM with a purely FNN architecture, because policies correspond to short token sequences (typically $L \le 20$), we concatenate all tokens into a single vector and process them through a FNN. Third, we omit both deep supervision and the ACT halting mechanism; the short sequence length and moderate recursion depth ($T \cdot m \approx 16$ to $64$ total steps) permit full gradient flow through the entire unrolled computation without memory or stability issues.

The remainder of this section details the construction of the input sequence, Section~\ref{subsec:tokens}, the recursive core that implements equations~\eqref{eq:z_update}--\eqref{eq:y_update},  Section~\ref{subsec:recursive_core}, and the output head that produces the final prediction, Section~\ref{subsec:metrics}.

\subsection{Covariate Tokens and Prefix Latent Tokens}
\label{subsec:tokens}

For each policy $i=1,\ldots, n$ we assume that the continuous and categorical feature vectors $(\bx_i,\bc_i)$ are transformed into $L = p_r + p_c$ embeddings
\[
\be_{i,1}, \dots, \be_{i,L} \in \R^d,
\]
using a feature-specific encoding pipeline; for the French MTPL portfolio this pipeline is described in Section~\ref{sec:example_feature_encoding}, and for more details we also refer to \citet{richman2025credibilitytransformer}. 

Following the Tab-TRM philosophy, we introduce two additional \emph{prefix tokens} representing the current answer and the latent reasoning state. Concretely, we maintain two trainable vectors
\begin{equation*}
\ba^{(0)} \in \R^{d_a},
\qquad
\bz^{(0)} \in \R^{d_z},
\end{equation*}
which are initialized the same across all policies. 
These are tiled across the entire sample and prepended to the feature tokens, yielding the initial sequence, for policies $i=1,\ldots, n$,
\begin{equation}
\bS_i^{(0)} = \bigl[\, \ba^{(0)}, \bz^{(0)}, \be_{i,1}, \dots, \be_{i,L} \,\bigr].
\label{eq:initial_sequence_tabtrm}
\end{equation}
Note that the prefix tokens may have different dimensions $d_a$ and $d_z$ than the feature tokens $d$; in our implementation we project all tokens to a common working dimension before flattening. We do not use positional encodings because each token position corresponds to a fixed, named tariff factor (covariate component); the token ordering is fixed and acts as an implicit feature identity. This keeps the architecture as simple as possible: the only structural distinction is between the two prefix tokens and the $L$ feature tokens.

For clarity, Table~\ref{tab:dimensions} summarises the key dimensions used in the French MTPL application  in Section~\ref{sec:example}, below.

\begin{table}[htbp]
\centering
\caption{Summary of key dimensions for the French MTPL application.}
\label{tab:dimensions}
\begin{tabular}{lcc}
\toprule
\textbf{Symbol} & \textbf{Description} & \textbf{Example Value} \\
\midrule
$L$ & Number of feature tokens & 9 \\
$d=d_a=d_z$ & Token embedding dimension & 28 \\
$L+2$ & Total sequence length (incl.\ $\ba,\bz$) & 11 \\
$(L+2)d$ & Flattened input dimension to $f_z$ & 308 \\
\bottomrule
\end{tabular}
\end{table}

\subsection{Tiny Recursive Reasoning Core}
\label{subsec:recursive_core}

The core of Tab-TRM is a small recursive network that repeatedly refines the latent scratchpad $\bz$ and the answer token $\ba$, reusing the same parameters at each recursion step. This mirrors the TRM idea of a tiny network that learns an ``improvement operator'' for $(\ba,\bz)$, rather than a deep stack of separate layers.

Let $\bS_i^{(t)}$ denote the token sequence after $t=1,\ldots, T$ \emph{outer} iterations
\begin{equation}
\bS_i^{(t)} = \bigl[\, \ba_i^{(t)}, \bz_i^{(t)}, \be_{i,1}, \dots, \be_{i,L} \,\bigr],
\label{eq:initial_sequence_tabtrm tt}
\end{equation}
with $\ba_i^{(t)} \in \R^{d_a}$ and $\bz_i^{(t)}  \in \R^{d_z}$ being
the first two tokens (answer and latent state). Each outer iteration consists of two stages: an inner recursion that updates $\bz_i^{(t)}$ $m$ times given the current sequence, and a single answer update that refines $\ba_i^{(t)}$, see \eqref{eq:z_update}-\eqref{eq:y_update}. Note that only in the initial token \eqref{eq:initial_sequence_tabtrm} the answer and the reasoning token are not instance dependent.

We use two small FNNs for these updates: a network $f_z$ for the latent recursion, and a network $f_a$ for the answer update. The depth of each FNN is a hyper-parameter (0--5 hidden layers) with GELU activation function; in the Optuna-selected configuration that we will present later both FNNs have zero hidden layers, yielding a single affine map followed by a GELU nonlinear activation. All parameters in $f_z$ and $f_a$ are shared across all inner and outer iterations.

\subsubsection*{Stage 1: Latent Recursion on $\bz$}

Given $\bS_i^{(t)}$, we initialise
\[
\bS_i^{(t,0)} = \bS_i^{(t)}, \qquad \bz_i^{(t,0)} = \bz_i^{(t)}.
\]
For $s = 0,\dots,m-1$ we perform a residual update of $\bz_i^{(t)}$:
\begin{align}
\bu_i^{(t,s)} &= \Flatten\bigl( \bS_i^{(t,s)} \bigr) \in \R^{d_a+d_z+L d}, \\
\tilde{\bu}_i^{(t,s)} &= \LayerNorm\bigl( \bu_i^{(t,s)} \bigr), \\\label{what about i?}
\Delta \bz_i^{(t,s+1)} &= f_z\bigl( \tilde{\bu}_i^{(t,s)} \bigr), \\
\bz_i^{(t,s+1)} &= \bz_i^{(t,s)} + \Delta \bz_i^{(t,s+1)}, \\
\bS_i^{(t,s+1)} &= \bS_i^{(t,s)} \text{ with token 1 replaced by } \bz_i^{(t,s+1)}.
\end{align}
After $m$ inner steps we write $\bz_i^{(t+1)}:=\bz_i^{(t,m)}$.
This stage corresponds to the TRM latent recursion, where $\bz$ is repeatedly updated from $(\ba,\bz, \bx, \bc)$, except that here the ``input'' $(\bx, \bc)$ is the short sequence of feature tokens and the update is implemented by a tiny FNN rather than a Transformer block.

\subsubsection*{Stage 2: Answer Update from $(\ba,\bz)$}

We next refine the answer token using the updated latent state. We form
\begin{align}
\tilde{\bS}_i^{(t)} &= \LayerNorm\left(\bigl[\, \ba_i^{(t)}, \bz_i^{(t+1)}, \be_{i,1}, \dots, \be_{i,L} \,\bigr]\right), \\
\label{tilde : not}
\tilde{\ba}_i^{(t)} &= \tilde{\bS}^{(t)}_{i}[0],
\qquad
\tilde{\bz}_i^{(t+1)} = \tilde{\bS}^{(t)}_{i}[1], \\
\Delta \ba_i^{(t+1)} &= f_a\bigl( \operatorname{Concat}\bigl(\tilde{\ba}_i^{(t)}, \tilde{\bz}_i^{(t+1)}\bigr) \bigr), \\
\ba_i^{(t+1)} &= \ba_i^{(t)} + \Delta \ba_i^{(t+1)},
\end{align}
where in \eqref{tilde : not}, we select tokens 0 and 1 of $\tilde{\bS}_i^{(t)}$ reflecting the answer and the reasoning tokens.
We then assemble the next outer sequence as
\begin{equation}
\bS_i^{(t+1)} = \bigl[\, \ba_i^{(t+1)}, \bz_i^{(t+1)}, \be_{i,1}, \dots, \be_{i,L} \,\bigr].
\end{equation}

\subsubsection*{Outer Recursion and Effective Depth}

The outer recursion is repeated $T$ times,
\[
\bS_i^{(0)} \;\mapsto\; \bS_i^{(1)} \;\mapsto\; \cdots \;\mapsto\; \bS_i^{(T)}= \bigl[\, \ba_i^{(T)}, \bz_i^{(T)}, \be_{i,1}, \dots, \be_{i,L} \,\bigr],
\]
yielding a final answer token $\ba_i^{(T)}$.

Let $n_{\text{layers}}$ denote the number of affine layers inside $f_z$ and $f_a$ (two in our experiments). A single forward pass through Tab-TRM has an effective depth of order
\[
n_{\text{eff}} \approx n_{\text{layers}} \, m  \, T,
\]
even though the number of \emph{distinct} parameters is tiny. For a typical configuration with $m=6$ and $T=3$, the effective depth is comparable to a $40$ to $50$ layer FNN, but with far fewer parameters and a much stronger inductive bias: the network is forced to implement a general-purpose ``improvement operator'' that can be applied repeatedly.

In contrast to HRM and TRM on grid-based puzzles, where deep supervision and ACT are used to manage memory and training cost, we simply backpropagate through all $mT$ latent updates and $T$ answer updates. This is feasible here because the sequence length $L+2$ is small and the per-step networks are tiny. Conceptually, our recursion corresponds to one TRM supervision step with all updates unrolled.

\subsection{Decoder, Poisson Deviance Loss and Metrics}
\label{subsec:metrics}

We interpret the final answer tokens $\ba_i^{(T)}$ as a compact representation of the policy's risk profile. 
We then select a small FNN $f_o:\R^{d_a} \to \R$ to map the answer token to the prediction. Using the exponential output activation, we set for the estimated expected frequency
\begin{equation}
F_\theta(\bx_i, \bc_i) = \exp\left(f_o(\ba^{(T)}_i)\right),
\label{eq:lambda_mu}
\end{equation}
which ensures compatibility with the Poisson GLM form~\eqref{eq:poisson_model}. For the network
$f_o$ we select a FNN of depth 2 and GELU activations; for a detailed exposition of FNN we refer to Chapter 5 of  \cite{wuethrich2025aitools}.


This gives us a parametrized class $\{F_\theta\}_\theta$ of regression functions, where the parameter $\theta$ collects all the token embedding parameters of $\bx$ and $\bc$ and the network parameters of the three networks $f_z$, $f_a$ and $f_o$. We train these weights by maximizing the Poisson log-likelihood which is equivalent to minimizing the Poisson deviance loss, see Chapter 2 of  \cite{wuethrich2025aitools}.
The Poisson deviance loss is given by
%
%
%
\begin{equation}
{\cal L}(\theta; \mathcal{D}) = \frac{2}{|\mathcal{D}|} \sum_{i \in \mathcal{D}} \left[ v_i F_\theta(\bx_i, \bc_i) - y_i - y_i \log \left(\frac{v_i F_\theta(\bx_i, \bc_i)}{y_i }\right) \right],
\label{eq:poisson_deviance}
\end{equation}
we refer to Table 2.2 in \cite{wuethrich2025aitools}. We use this Poisson deviance loss \eqref{eq:poisson_deviance} to train the model on a training sample $\mathcal{D}_{\rm train}$ to determine $\widehat{\theta}$,  and it is used for model validation on an independent hold-out test sample $\mathcal{D}_{\rm test}$.

\subsection{Optimization and Hyper-parameter Search}
\label{subsec:optuna}

We implement Tab-TRM in Keras and train it with the AdamW optimizer (decoupled weight decay), with learning rate, weight decay, Adam $\beta_2$, and dropout rates treated as hyper-parameters. Mild $\ell_1$--$\ell_2$ penalties are applied to the continuous encoding weights and categorical embeddings; when searching through the hyper-parameters we found that setting the value of the $\ell_1$ and $\ell_2$ penalties to the same value yielded the best results. Training is performed with mini-batches of size $4{,}096$, a 10\% validation split, early stopping on validation loss, and we reduce the learning rate by a factor of $0.5$ when the validation loss does not improve for $5$ epochs.

The architecture is controlled by a small set of interpretable hyper-parameters. These include the embedding dimension $d$ (shared across all tokens, typically not exceeding 60); the inner recursion depth $m$ and the number of outer iterations $T$; the numbers and widths of hidden layers in $f_z$, $f_a$ and $f_o$, dropout rates and regularization strengths; and the AdamW learning rate and $\beta_2$ parameter. We performed global hyper-parameter optimization with Optuna \citep{optuna}. For each trial we sample a configuration within pre-specified bounds, instantiate Tab-TRM with this configuration, train on $\mathcal{D}_{\mathrm{train}}$ with early stopping on a 10\% validation subset, reload the best checkpoint (as measured by the validation loss), and report the \emph{validation} Poisson deviance loss as the trial objective. After selecting the hyper-parameters, we evaluate the final model on the hold-out test sample and report the test Poisson deviance loss once. The search space and bounds are summarized in Appendix~\ref{app:hyperparams}.

\section{Interpreting Tab-TRM}
\label{sec:interpretation}

The recursive structure of Tab-TRM admits several complementary interpretations that connect to well-established modeling frameworks. In this section we first show that Tab-TRM can be viewed as a discrete-time state-space model, Section~\ref{subsec:state_space}, and secondly draw a connection to gradient boosting through a stagewise-additive lens, Section~\ref{subsec:recurrent_boosting}. These perspectives provide insight into the model's behavior and facilitate comparison with classical actuarial and machine learning methods.

\subsection{Connection to State-Space Models}
\label{subsec:state_space}

Tab-TRM can be viewed as a deterministic state-space system evolving in a learned latent space. For a fixed policy, the feature tokens $\be_1,\dots,\be_L$ are constant, so the recursion acts only on the latent state $(\ba,\bz)$. It is therefore natural to define the state
\[
\bs^{(t)} \;=\; \begin{pmatrix}\ba^{(t)}\\ \bz^{(t)}\end{pmatrix} \in \R^{d_a + d_z},
\]
and to interpret each outer recursion step as a discrete ``time'' index. The inner loop updates $\bz^{(t)}$ $m$ times using the current answer token and the fixed features, yielding $\bz^{(t+1)}$, and then the outer update refines $\ba^{(t)}$. This defines a deterministic nonlinear transition map of the form
\begin{equation}
\bs^{(t+1)} \;=\; \mathcal{F}_\theta\left(\bs^{(t)};\be_1,\dots,\be_L\right),
\label{eq:state_transition}
\end{equation}
where $\mathcal{F}_\theta$ is the composition of the $m$ inner updates of $f_z$ and the single outer update of $f_a$, see  \eqref{eq:z_update}-\eqref{eq:y_update}. The observation equation is the decoder applied to the answer token,
\begin{equation}
{F}_{\theta}^{(t)}(\bx_i, \bc_i) = \exp\left(f_o(\ba_i^{(t)})\right),
\label{eq:state_observation}
\end{equation}
with $f_o(\cdot)$ being the decoder FNN, see \eqref{eq:lambda_mu}. Thus, the full Tab-TRM is a nonlinear state-space model in a learned basis, with constant inputs given by the embedded covariates.

\paragraph{From nonlinear to linear recursion.}
In the full Tab-TRM, the update maps $f_z$ and $f_a$ are small FNNs with nonlinear activations (GELU in our implementation). For example, a two-layer update has the form
\[
f_z(\mathbf{u}) = \bW^{(2)}_z \,\phi\left(\bW^{(1)}_z \mathbf{u} + \bb^{(1)}_z\right) + \bb^{(2)}_z,
\quad
f_a(\mathbf{v}) = \bW^{(2)}_a \,\phi\left(\bW^{(1)}_a \mathbf{v} + \bb^{(1)}_a\right) + \bb^{(2)}_a,
\]
so the transition $\mathcal{F}_\theta$ is nonlinear because of the nonlinearity of $\phi(\cdot)$. The linear modification is obtained by \emph{removing the activations} (i.e., setting $\phi$ to the identity) and, equivalently, using zero hidden layers (a single affine map) for both $f_z$ and $f_a$. In that case
\[
f_z(\mathbf{u}) = \bW_z \mathbf{u} + \bb_z,
\qquad
f_a(\mathbf{v}) = \bW_a \mathbf{v} + \bb_a,
\]
so the recursion becomes exactly linear in $(\ba,\bz)$ and in the feature tokens. With residual connections this remains a linear update, and if LayerNorm is disabled the state transition is strictly linear. If LayerNorm is retained, the dynamics are a linear map followed by normalization (piecewise-linear), so the state-space form remains a close local approximation. The observation equation remains the same exponential decoder, so the overall model is linear in state evolution with a GLM-style nonlinear readout.

\paragraph{Linear reformulation (exact state-space form).}
Let $\bar{\be} = \frac{1}{L}\sum_{\ell=1}^L \be_\ell$ denote the mean feature token. For expositional clarity, we present the linear reformulation using this averaged representation; the actual implementation uses the full concatenated vector $[\be_1,\ldots,\be_L]$ with position-specific weight blocks, which is equivalent to the mean-token form under tied weights across positions. We write the linear updates as
\[
\Delta\bz = \bW_{zz}\bz + \bW_{az}\ba + \bW_f \bar{\be} + \bb_z,
\qquad
\Delta\ba = \bW_{aa}\ba + \bW_{za}\bz + \bb_a,
\]
with residual connections. Define $\bA_z = \bI + \bW_{zz}$ and the inner-loop sum
\[
\bS_{\mathrm{sum}} = \sum_{k=0}^{m-1} \bA_z^k.
\]
After $m$ inner updates, the reasoning state is
\begin{equation}
\bz^{(t+1)}
= \bA_z^m \bz^{(t)} + \bS_{\mathrm{sum}} \bigl(\bW_{az}\ba^{(t)} + \bW_f \bar{\be} + \bb_z\bigr).
\label{eq:z_star_linear}
\end{equation}
The outer update gives
\begin{equation}
\ba^{(t+1)}
= \bigl(\bI+\bW_{aa}\bigr)\ba^{(t)} + \bW_{za}\bz^{(t+1)} + \bb_a.
\label{eq:y_update_linear}
\end{equation}
Stacking these yields an exact linear state-space system
\begin{equation}
\bs^{(t+1)} = \bA\,\bs^{(t)} + \bB\,\bar{\be} + \bc,
\label{eq:linear_state_space}
\end{equation}
with block matrices
\[
\bA =
\begin{bmatrix}
\bI+\bW_{aa} + \bW_{za}\bS_{\mathrm{sum}}\bW_{az} & \bW_{za}\bA_z^m \\[0.2em]
\bS_{\mathrm{sum}}\bW_{az} & \bA_z^m
\end{bmatrix},
~
\bB =
\begin{bmatrix}
\bW_{za}\bS_{\mathrm{sum}}\bW_f \\[0.2em]
\bS_{\mathrm{sum}}\bW_f
\end{bmatrix},
~
\bc =
\begin{bmatrix}
\bb_a + \bW_{za}\bS_{\mathrm{sum}}\bb_z \\[0.2em]
\bS_{\mathrm{sum}}\bb_z
\end{bmatrix}.
\]
Equation~\eqref{eq:linear_state_space} is an exact deterministic linear state-space model with constant input $\bar{\be}$, followed by the same decoder as in the nonlinear case. We will come back to this in Section \ref{Linear Update Architecture}, below.

\subsection{A Stagewise-Additive View: Tab-TRM as Recurrent Boosting}
\label{subsec:recurrent_boosting}

Gradient boosting \citep{friedman2001gbm} constructs an additive predictor of the form
\[
F_M(x) = F_0(x) + \sum_{m=1}^M \nu_m h_m(x),
\]
where each weak learner $h_m$ is fitted to (pseudo-)residuals under the current model and $\nu_m$ is a shrinkage coefficient.
Tab-TRM admits a closely related \emph{stagewise correction} interpretation. Let
\[
\eta_i^{(t)} = f_o \left(\ba_i^{(t)}\right) = \log F^{(t)}_\theta(\bx_i,\bc_i)
\]
denote the model's log-prediction at outer iteration $t$, where $f_o:\R^{d_a}\to\R$ is the decoder mapping the answer token to a scalar. The outer recursion updates $\ba$ via residual increments:
\[
\ba_i^{(t+1)} = \ba_i^{(t)} + f_a\left(\operatorname{Concat}\left(\tilde{\ba}_i^{(t)},\tilde{\bz}_i^{(t+1)}\right)\right).
\]
If $f_o$ is linear (or locally linear), then $\eta^{(t)}$ evolves approximately additively:
\[
\eta_i^{(t+1)} \approx \eta_i^{(t)} + \underbrace{\left\langle \nabla f_o(\ba_i^{(t)}),\, f_a\left(\operatorname{Concat}\left(\tilde{\ba}_i^{(t)},\tilde{\bz}_i^{(t+1)}\right)\right)\right\rangle}_{=\Delta \eta_i^{(t)}},
\]
so that after $T$ outer steps the log-prediction decomposes as
\[
\eta_i^{(T)} \approx \eta_i^{(0)} + \sum_{t=0}^{T-1} \Delta \eta_i^{(t)}.
\]
This structure resembles boosting in that the predictor is refined by repeated additive corrections. However, there is a key architectural difference: whereas standard gradient boosting fits a \emph{new} weak learner at each stage, Tab-TRM reuses the \emph{same} small networks $f_z$ and $f_a$ across all refinement steps. Computational depth therefore increases without introducing additional parameter sets, yielding a form of \emph{recurrent boosting} where the correction operator is learned once and applied repeatedly.

\section{Application: French MTPL Data}
\label{sec:example}

We now demonstrate Tab-TRM on the benchmark French MTPL data of \cite{Dutang}; we use the cleaned data of \cite{wuethrichmerz2023statistical}. The goal is to model the claim frequency using the methodology of Section~\ref{sec:methodology}, together with an insurance-specific feature encoding pipeline.

\subsection{Data and Baseline Formulation}

Each observation corresponds to a single exposure period and contains several components. The response variable $Y_i \in \{0,1,2,\dots\}$ records the number of reported claims, while $v_i > 0$ denotes the exposure. The available covariates are split into two groups: a vector $\bx_i = (x_{i,1},\dots,x_{i,p_r})$ of $p_r = 5$ continuous tariff variables and a vector $\bc_i = (c_{i,1},\dots,c_{i,p_c})$ of $p_c = 4$ categorical tariff variables. In addition, a split indicator assigns each observation either to a training set or to a test set. We focus on the claim counts variable and calibrate a Poisson model
\begin{equation}\label{Poisson mu}
Y_i \mid \bx_i, \bc_i, v_i \sim \mathrm{Poisson}(\mu_i),
\qquad \text{ with mean }
\mu_i = v_i  F_\theta(\bc_i,\bx_i).
\end{equation}
As a simple reference we consider the null Poisson baseline on the training set, not considering any covariates, having empirical expected frequency estimate
\begin{equation}
\widehat{\lambda}_{\mathrm{null}} = \frac{\sum_{i \in \mathcal{D}_{\mathrm{train}}} Y_i}{\sum_{i \in \mathcal{D}_{\mathrm{train}}} v_i},
\end{equation}
which coincides with the intercept-only model of a Poisson GLM with $\log v_i$ as offset.

\subsection{Feature Transformations, Embeddings and Tokens}
\label{sec:example_feature_encoding}

For the French MTPL dataset we make the encoding pipeline of Section~\ref{subsec:tokens} concrete. The aim is to construct $L = p_r + p_c = 9$ feature tokens per policy, each living in $\R^d$, that serve as the Tab-TRM input sequence.

\subsubsection*{Continuous tariff variables}

For each continuous component $(x_{i,j})_{j=1}^{p_r}$ we first apply a robust, smooth clipping transformation, following the quantile-based preprocessing of \citet{holzmuller2024realmlp}. On the training set we compute empirical quantiles
\begin{equation}\label{quantiles range}
q_{j,0} < q_{j,1} < q_{j,2} < q_{j,3} < q_{j,4},
\end{equation}
corresponding to the 0th, 25th, 50th, 75th and 100th percentiles of $\{x_{i,j} : i \in \mathcal{D}_{\mathrm{train}}\}$. A positive scale $s_j$ is derived from the spread of the distribution (we use an inter-quantile range based on \eqref{quantiles range}, with safeguards against degeneracies), and we define
\begin{equation}
z_{i,j} = s_j \bigl( x_{i,j} - q_{j,2} \bigr),
\qquad
\tilde{x}_{i,j} = \frac{z_{i,j}}{\sqrt{1 + (z_{i,j}/3)^2}}.
\label{eq:clipped_transform_example}
\end{equation}
Near the median $q_{j,2}$ this transform is approximately linear, while for large positive or negative deviations it saturates smoothly towards values in $(-3,3)$, limiting the influence of outliers while preserving rank information. The identical $(q_{j,2},s_j)$ statistics are reused at prediction time.

To allow nonlinear effects of continuous factors while keeping the model low-capacity, we apply a learnable piecewise-linear encoding (PLE), inspired by \citet{gorishniy2022embeddings} and adapted to actuarial data following \citet{richman2025credibilitytransformer}. For each normalized feature $\tilde{x}_{i,j}$ we compute, on the pooled train normalized data, a grid of empirical deciles
\[
d_{j,0} < d_{j,1} < \ldots < d_{j,K}, \qquad K = 10,
\]
and initialize a custom Keras layer that maintains log bin-widths whose exponentials yield positive widths, accumulates these widths (plus a trainable start value) to form a monotone set of bin boundaries $b_{j,0},\dots,b_{j,K}$ and maps $\tilde{x}_{i,j}$ to a $(K+1)$-dimensional vector $\psi_j(\tilde{x}_{i,j})$ of piecewise-linear basis functions that are localized between these boundaries. We then obtain a $d$-dimensional continuous embedding via
\begin{equation}
\mathbf{s}_{i,j} = \psi_j(\tilde{x}_{i,j}) \in \R^{K+1},
\qquad
\be_{i,j}^{\mathrm{cont}} = \Phi\left( \bW_j^{\mathrm{cont}} \mathbf{s}_{i,j} + \mathbf{b}_j^{\mathrm{cont}} \right) \in \R^{d},
\end{equation}
with $\bW_j^{\mathrm{cont}}$ and $\mathbf{b}_j^{\mathrm{cont}}$ trainable and $\Phi$ a GELU activation. Mild $\ell_1$--$\ell_2$ penalties are applied to these weights. From an actuarial point of view, this acts as a learned rating curve for each continuous rating factor, initialized at decile-based knots and allowed to adapt during training.

\subsubsection*{Categorical rating factors}

Each categorical rating factor $(c_{i,j})_{j=1}^{p_c}$ takes values in a finite set $\{0,\dots,M_j\}$ after label encoding. We associate an embedding matrix $\bE_j^{\mathrm{cat}} \in \R^{(M_j+1) \times d}$ and map
\begin{equation}
\be_{i,j}^{\mathrm{cat}} = \bE_j^{\mathrm{cat}}[c_{i,j}] \in \R^{d},
\end{equation}
corresponding to level $c_{i,j}$ taken by policy $i$, i.e., 
$\be_{i,j}^{\mathrm{cat}}$ is the $(c_{i,j}+1)$-st row of $\bE_j^{\mathrm{cat}}$.
We apply light $\ell_1$--$\ell_2$ regularization to encourage shrinkage of rarely used levels.

\subsubsection*{Token representation for Tab-TRM}

For each policy $i=1,\ldots, n$ we now have $p_r = 5$ continuous embeddings $\be_{i,1}^{\mathrm{cont}},\dots,\be_{i,p_r}^{\mathrm{cont}}$ and $p_c = 4$ categorical embeddings $\be_{i,1}^{\mathrm{cat}},\dots,\be_{i,p_c}^{\mathrm{cat}}$. We stack them into $L = 9$ feature tokens
\[
\bigl[\, \be_{i,1},\dots,\be_{i,L} \,\bigr]^\top
=\bigl[\, \be_{i,1}^{\mathrm{cont}},\dots,\be_{i,p_r}^{\mathrm{cont}}, \be_{i,1}^{\mathrm{cat}},\dots,\be_{i,p_c}^{\mathrm{cat}}
 \,\bigr]^\top \in \R^{L \times d}.
\]
These tokens are fed into Tab-TRM by prepending the learned answer and latent tokens as in~\eqref{eq:initial_sequence_tabtrm} and \eqref{eq:initial_sequence_tabtrm tt} for $t\ge 1$,
\[
\bS_i^{(0)} = \bigl[\, \ba^{(0)}, \bz^{(0)}, \be_{i,1},\dots,\be_{i,L} \,\bigr]
\qquad \text{ and }\qquad
\bS_i^{(t)} = \bigl[\, \ba_i^{(t)}, \bz_i^{(t)}, \be_{i,1},\dots,\be_{i,L} \,\bigr],
\]
by applying the recursive core of Section~\ref{subsec:recursive_core}. No positional encodings are used; the organisation follows the actuarial separation between tariff factors and the global latent states $(\ba,\bz)$.

\subsection{Training Setup and Evaluation on the French MTPL Portfolio}

We train Tab-TRM on the French MTPL data, using the same training-test partition as in \cite{wuethrich2025aitools}. We use the Poisson deviance loss \eqref{eq:poisson_deviance}, with the optimization and hyper-parameter search set-up as described in Section~\ref{subsec:optuna}. For each Optuna trial we sample a candidate set of architectural and optimization hyper-parameters (embedding dimension $d$, recursion depths $m$ and $T$, widths and depths of $f_z$ and $f_a$, dropout rates, regularization strengths, AdamW parameters), train the model with early stopping on a 10\% validation subset of the training data, reload the best checkpoint and compute the Poisson deviance loss \eqref{eq:poisson_deviance} on the hold-out test set. Since we use the identical training-test split as in many other studies, the out-of-sample Poisson deviance losses are directly comparable.

\section{Results}
\label{sec:results}

\subsection{Hyper-parameter Selection and Model Performance}

Hyper-parameters are selected via Optuna, using the validation Poisson deviance loss as the objective. We conducted the search in two stages. First, a broad search over the full hyper-parameter space (see Table~\ref{tab:hyperparams}) revealed that the best validation scores were consistently achieved with \emph{single-layer} networks for both $f_z$ and $f_a$ - that is, zero hidden layers in each update function. A second, focused Optuna run restricted to single-layer configurations then identified the final architecture, reported in Table~\ref{tab:optuna_params}. To reduce variance, we train a 10-run nagging ensemble \citep{richman2022nagging} - averaging predictions across independently initialised runs - which improves test deviance from $23.630 \times 10^{-2}$ (single run) to $23.589\times 10^{-2}$, with mean per-run deviance $23.666\times 10^{-2}$.

This finding is noteworthy: Tab-TRM achieves competitive performance using zero-hidden-layer networks for $f_z$ and $f_a$, that is, a single affine projection followed by a GELU activation. Even these minimal nonlinear operators, when composed recursively $m \times T$ times through the Tab-TRM core, are sufficient to learn complex risk structures. This supports the TRM philosophy that computational depth through parameter reuse can substitute for model width and architectural complexity.

\begin{table}[htbp]
\centering
\caption{Optuna-selected hyper-parameters (best validation deviance).}
\label{tab:optuna_params}
\begin{tabular}{ll}
\toprule
\textbf{Hyper-parameter} & \textbf{Value} \\
\midrule
Embedding dimension $d=d_a=d_z$ & 28 \\
Outer steps $T$ & 6 \\
Inner iterations $m$ & 3 \\
Output FNN layer 1 hidden units & 19 \\
Output FNN layer 2 hidden units & 124 \\
Dropout (FNN1) & 0.2821 \\
Dropout (FNN2) & 0.4991 \\
$\ell_1$--$\ell_2$ regularization & $2.2539\times10^{-5}$ \\
Learning rate & 0.0021755 \\
Weight decay & 0.0239601 \\
Adam $\beta_2$ & 0.9594 \\
\bottomrule
\end{tabular}
\end{table}

We compare the Tab-TRM results against a range of benchmark models from the actuarial deep learning literature. Table~\ref{tab:benchmark_comparison} reports in-sample and out-of-sample Poisson deviance losses (in units of $10^{-2}$) for classical models, FNNs, and recent transformer-based architectures. Benchmark results are taken from \citet{wuethrichmerz2023statistical}, \citet{brauer2024cafft} and \citet{richman2025credibilitytransformer,richman2025pins}.

\begin{table}[htbp]
\centering
{\small
\begin{tabular}{l||c|cc|cc}
\hline
& \# &\multicolumn{2}{c|}{In-sample}&\multicolumn{2}{c}{Out-of-sample}\\
Model & Param.&\multicolumn{2}{c|}{Poisson loss} & \multicolumn{2}{c}{Poisson loss} \\
\hline\hline
Null model (intercept-only) &1&25.213& &25.445&\\
Poisson GLM3 &50& 24.084 && 24.102&\\
Poisson GAM &  (66.7) &23.920&&23.956&\\\hline
Plain-vanilla FNN &792& 23.728 & ($\pm$ 0.026)& 23.819&($\pm$ 0.017)\\
Ensemble plain-vanilla FNN &792& 23.691 && 23.783&\\\hline
CAFFT & 27,133& 23.715 &($\pm$ 0.047) & 23.807 &($\pm$ 0.017)\\
Ensemble CAFFT & 27,133& 23.630 & & 23.726 &\\\hline
Credibility Transformer & 1,746& 23.641 &($\pm$ 0.053) & 23.788 &($\pm$ 0.040)\\
Ensemble Credibility Transformer & 1,746& 23.562 && 23.711 &
\\ \hline
Tree-like PIN& 4,147& 23.593 &($\pm$ 0.046) & 23.740 &($\pm$ 0.025)\\
Ensemble Tree-like PIN& 4,147& 23.522 && 23.667 &\\
\hline
\textcolor{blue}{Tab-TRM}& \textcolor{blue}{14,820}& \textcolor{blue}{23.570} & \textcolor{blue}{($\pm$ 0.027)} & \textcolor{blue}{23.666} & \textcolor{blue}{($\pm$ 0.027)}\\
\textcolor{blue}{Ensemble Tab-TRM}& \textcolor{blue}{14,820}& \textcolor{blue}{23.496} && \textcolor{blue}{23.589} &\\
\hline
\end{tabular}}
\caption{Number of parameters, in-sample and out-of-sample Poisson deviance losses (units are in $10^{-2}$). Benchmark models are taken from \citet{wuethrichmerz2023statistical}, \citet{brauer2024cafft}, \citet{richman2025credibilitytransformer, richman2025pins}. Tab-TRM results are for the Optuna-selected configurations.}
\label{tab:benchmark_comparison}
\end{table}

Tab-TRM achieves an out-of-sample Poisson deviance loss of $23.630 \times 10^{-2}$ (single run) and $23.589 \times 10^{-2}$ (nagging ensemble), which is competitive with the best network architectures in the table. Tab-TRM uses 14,820 trainable parameters, about 45\% fewer than CAFFT (27,133) and within an order of magnitude of other compact tabular architectures such as the Tree-like PIN (4,147) and Credibility Transformer (1,746). The architecture trades model size for computational depth through parameter reuse across recursion steps.

\subsection{Interpretability Analysis}
\label{sec:trial56_interpretability}

To understand how Tab-TRM refines its predictions, we analyze the learned representations and recursive dynamics on a random subset of 512 test policies. For each policy we extract the initial token sequence from the trained encoder, run the recursive layer for $T=6$ outer steps with $m=3$ inner iterations, and record the answer token $\ba$ and reasoning token $\bz$ after each outer step. Per-step predictions are computed by passing the current $\ba$ token and exposure through the decoder.

\subsubsection*{Token dynamics and convergence}

\begin{figure}[htbp]
\centering
\includegraphics[width=0.9\linewidth]{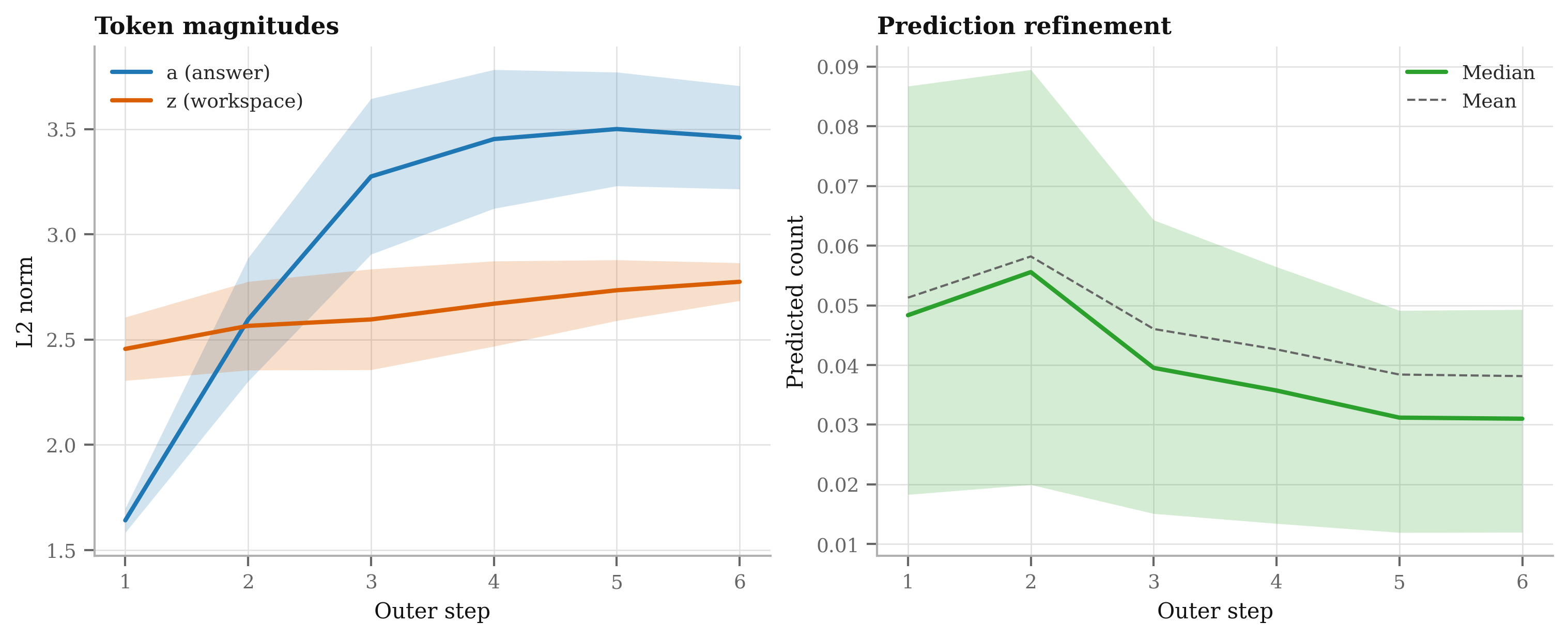}
\caption{Token magnitude and prediction refinement. Left: mean $\ell_2$ norms of $\ba$ and $\bz$ across outer steps with $\pm 1$ standard deviation bands over 512 random test policies. Right: per-step predictions computed from the $\ba$ token and exposure; line is the median, shaded band is the interquartile range, dashed line is the mean.}
\label{fig:trial56_token_dynamics}
\end{figure}

Figure~\ref{fig:trial56_token_dynamics} reports the evolution of token magnitudes and predictions across outer steps. The left panel shows the mean $\ell_2$ norm of $\ba$ and $\bz$ with $\pm 1$ standard deviation bands; the right panel displays the distribution of per-step predictions (median, interquartile range, and mean). Both token magnitudes and prediction distributions stabilize within a few steps, with most adjustment occurring early in the recursion. This rapid convergence is consistent with the TRM design: the recursive core learns an improvement operator that quickly refines an initial estimate rather than building up a prediction from scratch.

\begin{figure}[htbp]
\centering
\includegraphics[width=0.9\linewidth]{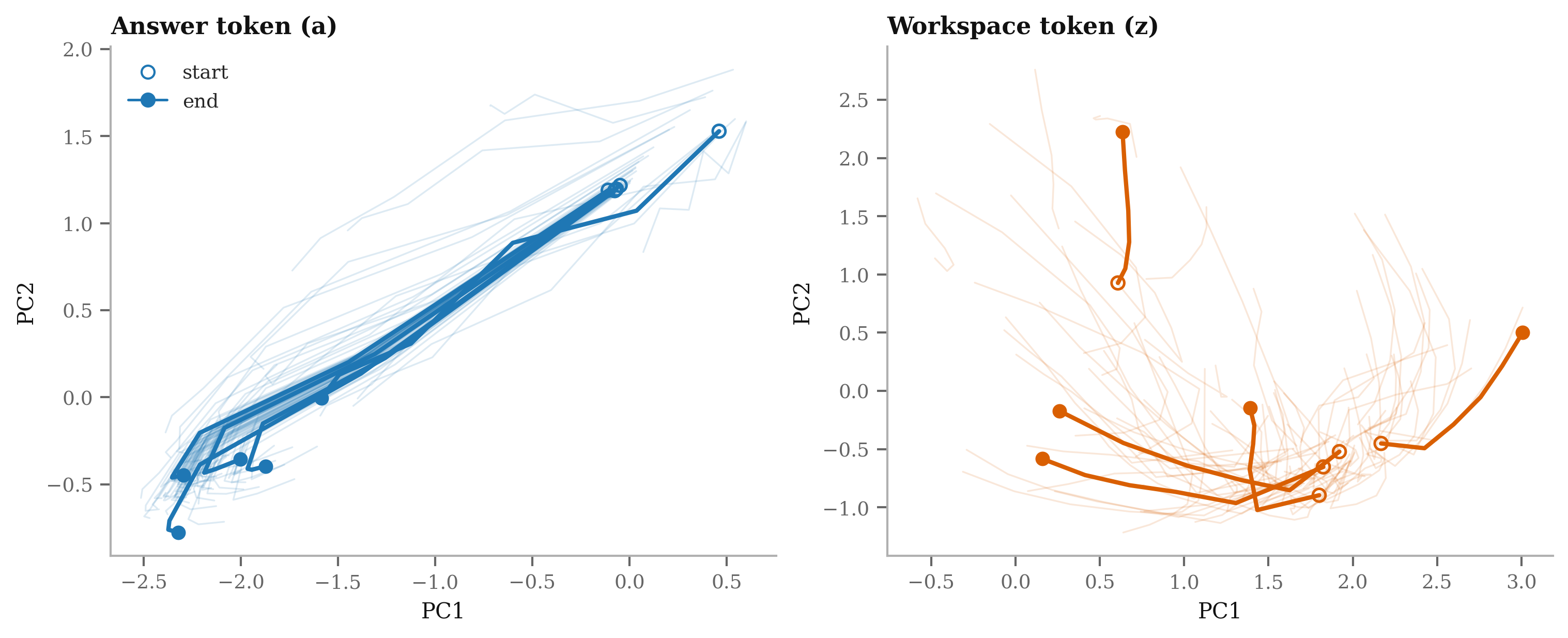}
\caption{Token trajectories in PCA space. PCA basis is fit to all $\ba$ and $\bz$ token vectors across outer steps for the 512-policy test subset. Faint lines show 80 random trajectories; bold lines highlight five policies selected by final prediction percentiles (5, 25, 50, 75, 95). Markers denote start (after the first outer step, open) and end (after the final outer step, filled) positions.}
\label{fig:trial56_pca_trajectories}
\end{figure}

Figure~\ref{fig:trial56_pca_trajectories} visualizes the token trajectories in a 2-dimensional PCA space, where the basis is fit to all $\ba$ and $\bz$ vectors across outer steps. The $\ba$ trajectories collapse toward a common region of the embedding space, while $\bz$ remains more dispersed across policies. This pattern supports the intended role of each token: $\bz$ acts as a flexible workspace that integrates policy-specific feature evidence, while $\ba$ converges to the final risk indication.

\subsubsection*{Feature alignment and attribution}

To understand how the model weights different covariates, we compute the cosine similarity (alignment) between each input feature embedding and the answer token at each outer step. Figure~\ref{fig:trial56_feature_attention} reports the mean alignment across the test subset, along with the change in alignment from the first to the final step for both $\ba$ and $\bz$. This reveals how the recursion progressively reweighs evidence across features, with some covariates gaining influence while others are downweighted.

Figure~\ref{fig:trial56_attribution} provides a complementary view through gradient-based attribution. For each policy we compute $\nabla_{\ba}\hat{\mu}$ from the decoder and project this gradient onto each input feature embedding; $\hat{\mu}$ reflects the estimated mean, see \eqref{Poisson mu}. The resulting attribution scores indicate which feature directions are most aligned with the final prediction, offering a saliency-style interpretation of the model's decision.

\subsubsection*{Update dynamics and linear structure}

\begin{figure}[htbp]
\centering
\includegraphics[width=0.9\linewidth]{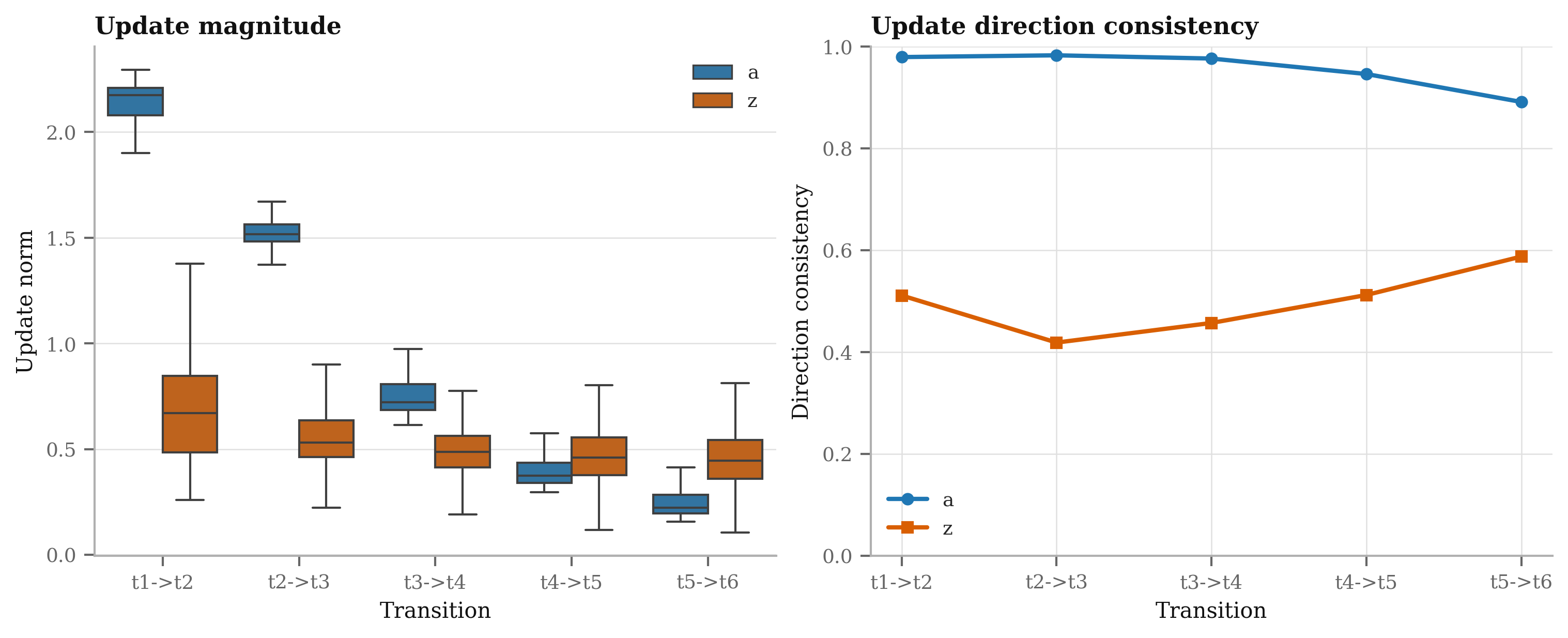}
\caption{Token update decomposition on the 512-policy test subset. Left: distribution of update magnitudes $\lVert \Delta \ba \rVert$ and $\lVert \Delta \bz \rVert$ by transition (differences between successive outer steps). Right: direction consistency defined as $\lVert \mathrm{mean}(\Delta/\lVert\Delta\rVert) \rVert$ across samples.}
\label{fig:trial56_updates}
\end{figure}

Figure~\ref{fig:trial56_updates} summarises the step-to-step changes $\Delta\ba$ and $\Delta\bz$ across transitions between successive outer steps. The left panel shows update magnitudes as boxplots; the right panel reports direction consistency, defined as the norm of the mean of unit-normalized update vectors. The $\ba$ updates are smaller and more directionally consistent than those of $\bz$, confirming that the answer token follows a stable refinement trajectory while the reasoning token explores more freely.

\begin{figure}[htbp]
\centering
\includegraphics[width=0.9\linewidth]{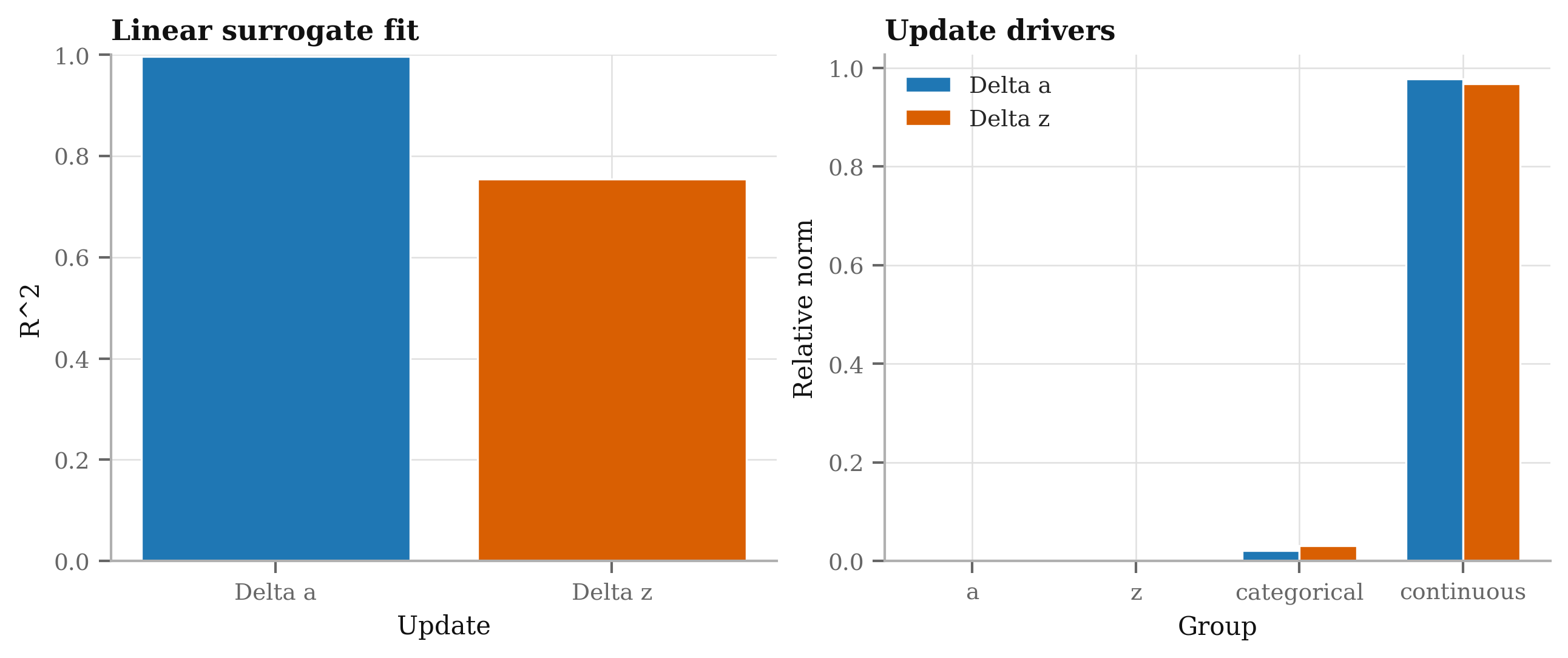}
\caption{Linear surrogate of update dynamics. Left: $R^2$ of least-squares fits predicting $\Delta \ba$ and $\Delta \bz$ from concatenated $[\ba^{(t)}, \bz^{(t)}, e_1,\dots,e_L]$. Right: relative coefficient norm by group (tokens vs categorical/continuous feature embeddings), showing which components contribute most to each update.}
\label{fig:trial56_linear_surrogate}
\end{figure}

To quantify how much of the recursive dynamics can be explained by linear operations, we fit least-squares surrogates of the form $\Delta\ba^{(t)} = A_t [\ba^{(t)}, \bz^{(t)}, \be_1,\ldots,\be_L]^\top + \mathbf{c}_t$, and similarly for $\bz^{(t)}$. Figure~\ref{fig:trial56_linear_surrogate} reports the $R^2$ of these fits (left) and the relative coefficient norms grouped by state tokens versus categorical and continuous feature embeddings (right). The high $R^2$ values indicate that the update dynamics are well-approximated by linear maps in the learned basis.

\begin{figure}[htbp]
\centering
\includegraphics[width=0.9\linewidth]{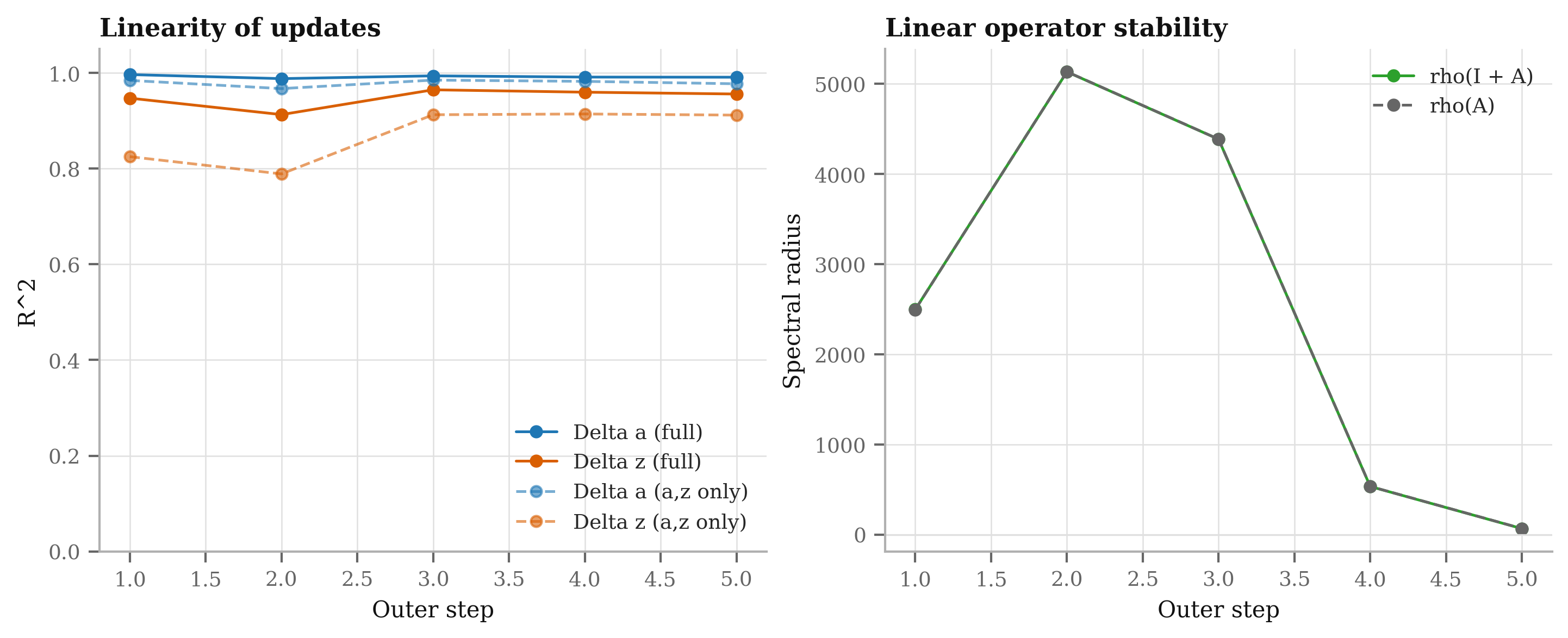}
\caption{Linearity diagnostics by recursion step. Left: per-step $R^2$ of linear fits for $\Delta \ba^{(t)}$ and $\Delta \bz^{(t)}$, shown for models using the full feature context and using $(\ba^{(t)},\bz^{(t)})$ only. Right: spectral radius of the effective linear update map $\bI + A_t$ (solid) and of the update operator $A_t$ itself (dashed), estimated from least-squares fits on the concatenated state.}
\label{fig:trial56_linear_diagnostics}
\end{figure}

Figure~\ref{fig:trial56_linear_diagnostics} extends this analysis by reporting per-step $R^2$ for models using the full context versus state-only inputs, along with the spectral radius of the estimated per-step update operator $A_t$. The spectral radii of the per-step updates remain moderate, indicating that individual refinement steps do not amplify state norms excessively. These diagnostics reinforce the state-space interpretation of Section~\ref{subsec:state_space}: Tab-TRM behaves as a near-linear dynamical system with a nonlinear observation equation. Note that Section~\ref{sec:linear_tabtrm} presents a complementary analysis for a fully linearized variant, where the effective outer-step transition matrix has spectral radius $1.44$, the dynamics are thus finite-step refinements rather than strict contractions to a fixed point.

\subsubsection*{Local coefficients}

\begin{figure}[htbp]
\centering
\includegraphics[width=0.9\linewidth]{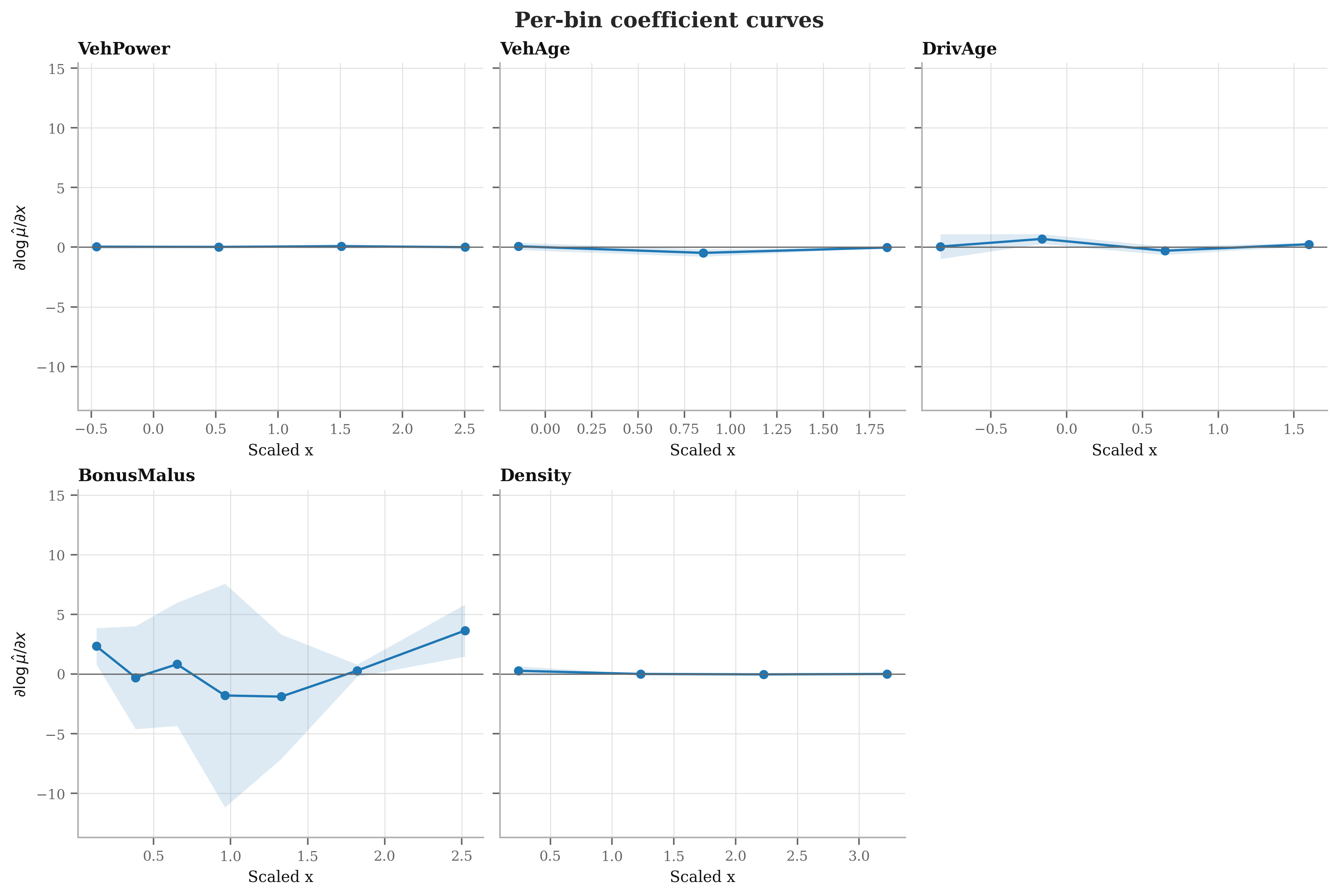}
\caption{Local coefficient analysis for categorical and continuous features. Per-bin averages of $\partial \log\hat{\mu} / \partial x_j$ using the learned piecewise-linear encoder bin edges, shown as coefficient curves for each continuous variable.}
\label{fig:trial56_local_coeffs_curves}
\end{figure}

Figures~\ref{fig:trial56_local_coeffs} and \ref{fig:trial56_local_coeffs_curves} report coefficient-style effects for categorical and continuous variables by differentiating the log-prediction $\log\hat{\mu}$ with respect to each scaled input. The former figure shows the mean derivative $\partial \log\hat{\mu} / \partial x_j$ with $\pm 1$ standard deviation across policies; the latter figure aggregates these derivatives within the learned piecewise-linear encoder bins to recover per-bin slope profiles for the continuous variables. These ``local coefficients'' provide an actuarial interpretation of the model's response to changes in each covariate.

\subsubsection*{Key insights}

The interpretability analyses yield two complementary sets of insights: one concerning the Tab-TRM architecture itself, and one concerning the risk structure of the French MTPL data.

\medskip

\emph{Insights about the model.} The recursive dynamics of Tab-TRM are remarkably well approximated by linear operations in the learned embedding space ($R^2 > 0.9$ for the linear surrogates). This validates the state-space interpretation developed in Section~\ref{subsec:state_space}, and it explains why even zero-hidden-layer update networks achieve competitive performance: the model's expressive power comes primarily from (i) the nonlinear feature encoders (piecewise-linear embeddings for continuous variables, entity embeddings for categorical variables), (ii) the recursive composition of updates, and (iii) the decoder. The moderate per-step spectral radii indicate stable refinement dynamics within the chosen recursion budget. The answer token $\ba$ follows a consistent refinement direction across policies, while the reasoning token $\bz$ acts as a flexible scratchpad that integrates heterogeneous feature evidence before distilling it into the final prediction.

\medskip

\begin{figure}[htbp]
\centering
\includegraphics[width=0.9\linewidth]{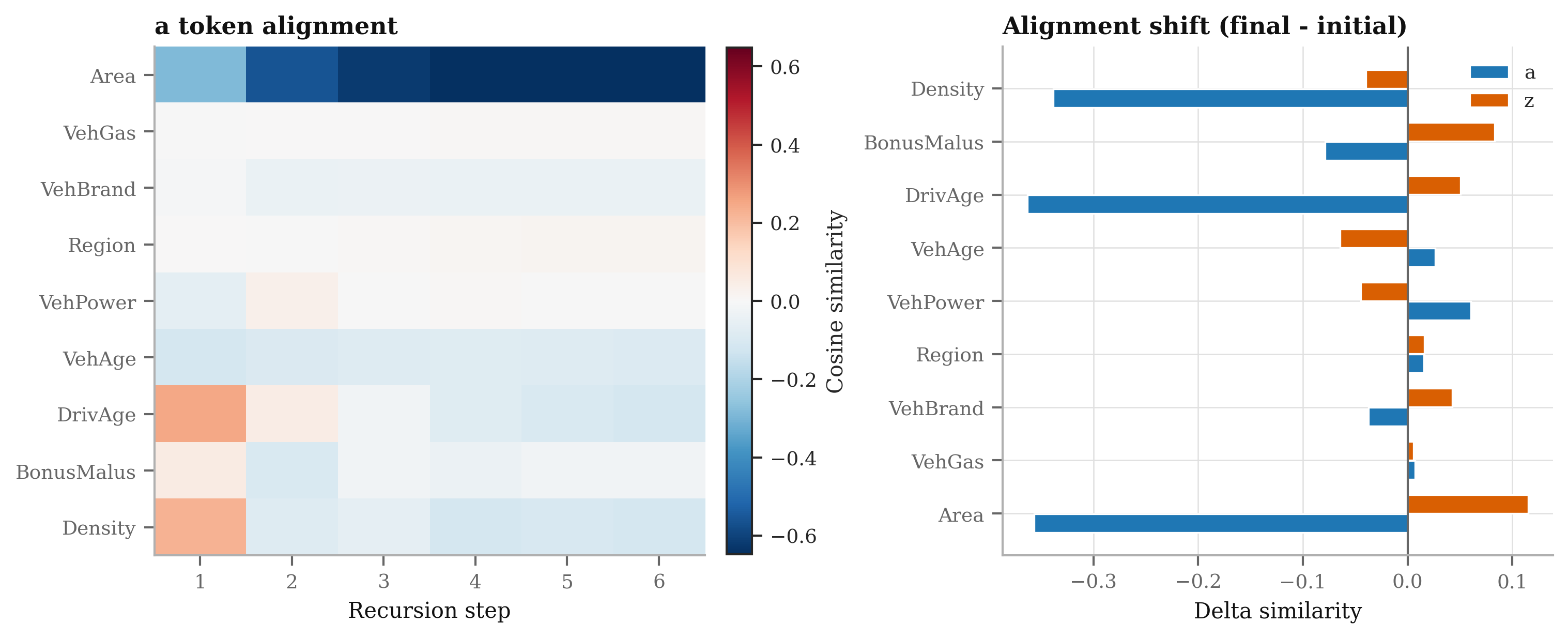}
\caption{Feature-to-token alignment. Left: mean cosine similarity between each feature embedding (encoder feature tokens) and the $\ba$ token at each outer step for the 512-policy test subset. Right: alignment shift (final minus initial) for both $\ba$ and $\bz$.}
\label{fig:trial56_feature_attention}
\end{figure}

\begin{figure}[htbp]
\centering
\includegraphics[width=0.8\linewidth]{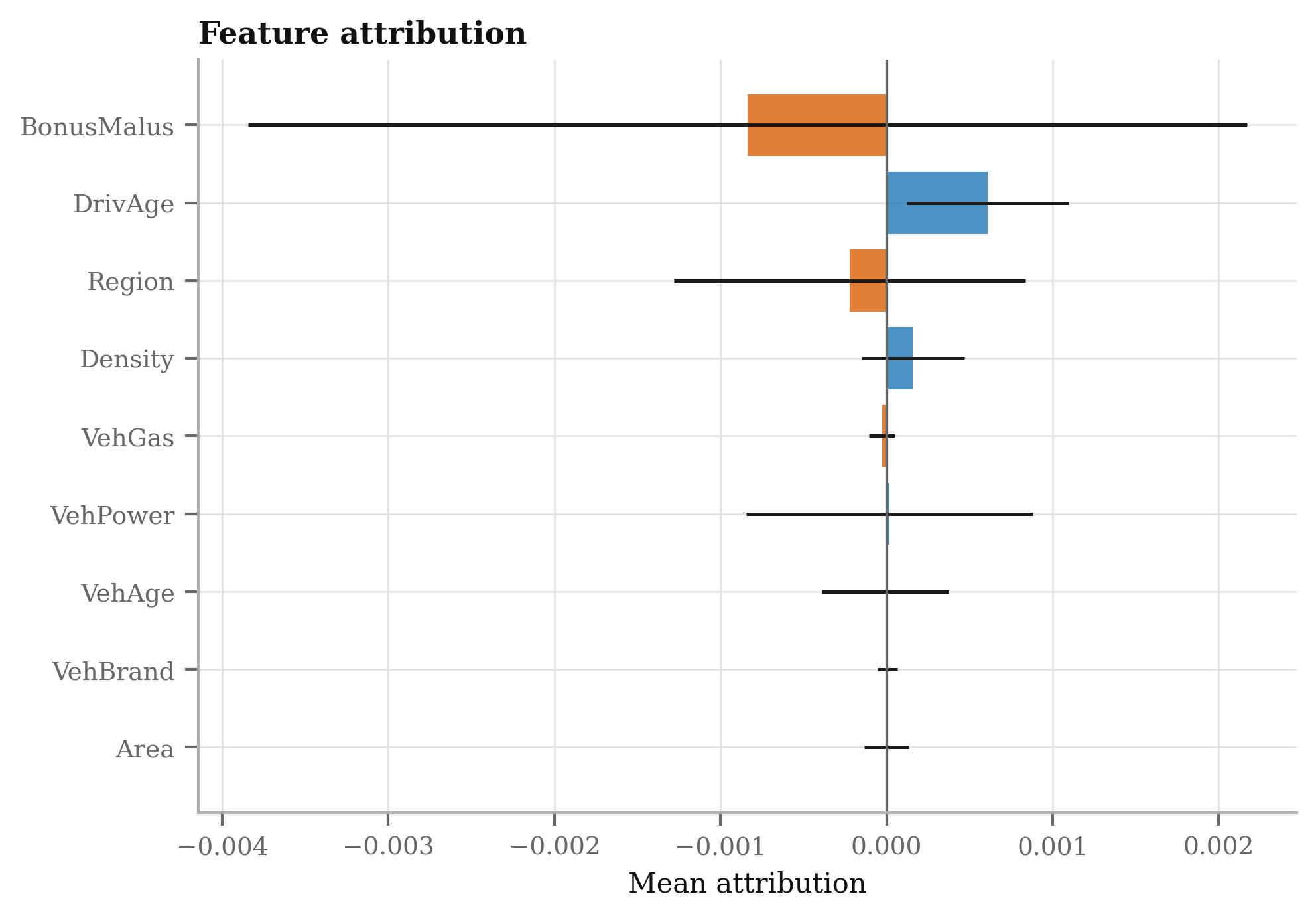}
\caption{Gradient-based feature attribution. For each policy, we compute $\nabla_{\ba} \hat{\mu}$ from the decoder and project it onto each encoder feature embedding; bars show mean attribution with $\pm$1 standard deviation across the 512-policy test subset.}
\label{fig:trial56_attribution}
\end{figure}

\begin{figure}[htbp]
\centering
\includegraphics[width=0.9\linewidth]{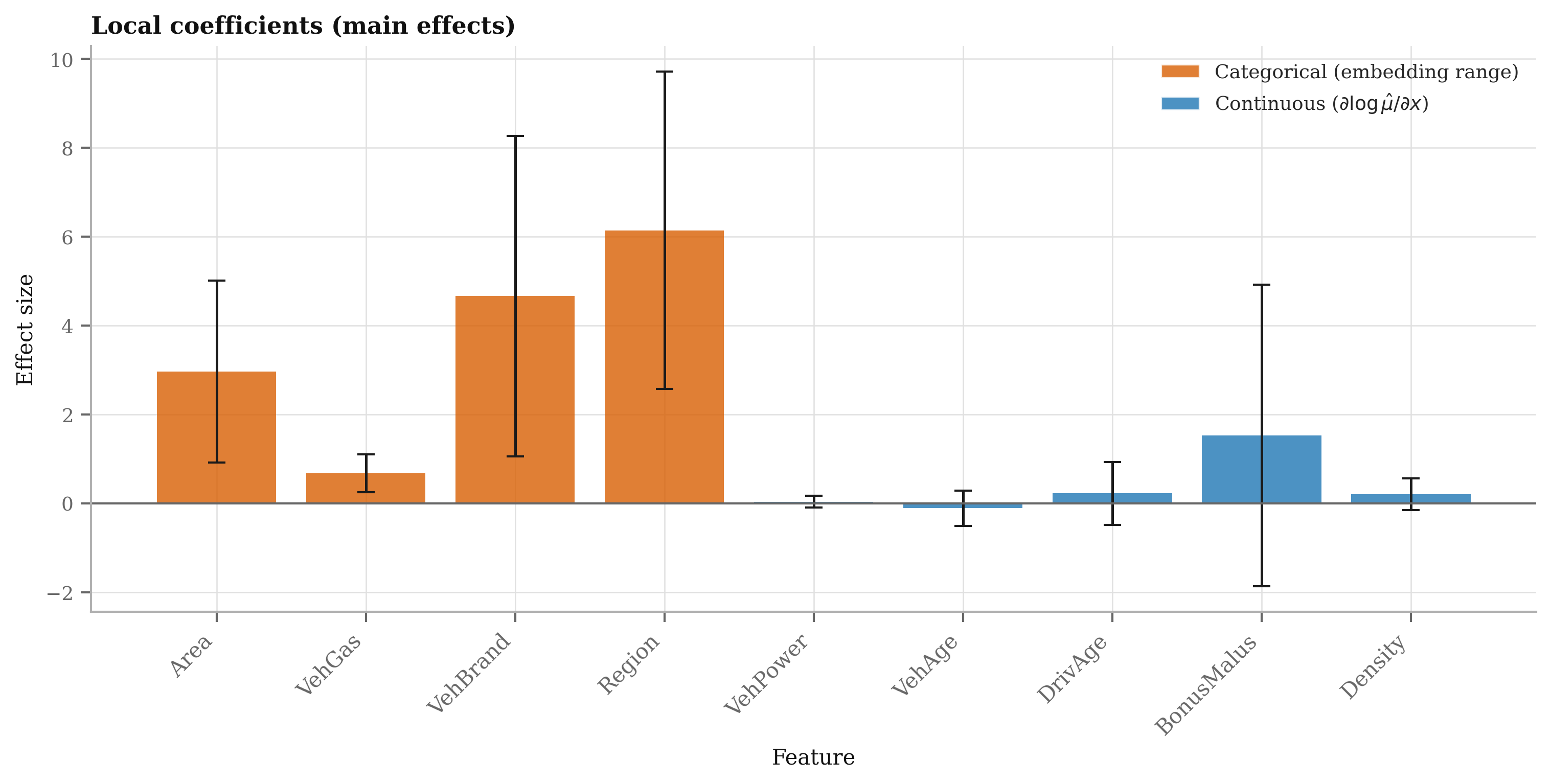}
\caption{Local coefficient analysis for categorical and continuous features. Mean $\partial \log\hat{\mu} / \partial x_j$ with $\pm$1 standard deviation across the 512-policy test subset (inputs use the scaled continuous variables from the data pipeline).}
\label{fig:trial56_local_coeffs}
\end{figure}

\emph{Insights about the French MTPL data.} The feature alignment and gradient attribution analyses (Figures~\ref{fig:trial56_feature_attention}--\ref{fig:trial56_attribution}) reveal which tariff factors most strongly influence the predicted claim frequency. Among the continuous variables, the bonus-malus level and the driver age show the largest alignment shifts and attribution scores, consistent with their well-documented importance in European motor pricing. Vehicle power and vehicle age also contribute meaningfully, while population density exhibits a more moderate effect. Among the categorical variables, region and vehicle brand emerge as the most influential, reflecting geographic risk variation and fleet-specific effects. The local coefficient curves in Figure~\ref{fig:trial56_local_coeffs} further reveal nonlinear rating structures: for instance, claim frequency decreases with driver age until middle age and then flattens, while the bonus-malus effect is approximately log-linear. These patterns align with standard actuarial intuition and with findings from classical GLM analyses of the same portfolio, providing confidence that Tab-TRM has learned economically meaningful risk relationships rather than spurious correlations.

\medskip

Taken together, these analyses reveal that Tab-TRM relies less on deep nonlinear composition and more on repeated, largely linear refinements. Early recursion steps move the prediction into the appropriate region; later steps apply smaller corrections. The reasoning token $\bz$ integrates feature evidence, while the answer token $\ba$ stabilizes into the final prediction.

\subsection{Test-time recursion selection}
\label{sec:trial56_testtime}

\begin{figure}[htbp]
\centering
\includegraphics[width=0.8\linewidth]{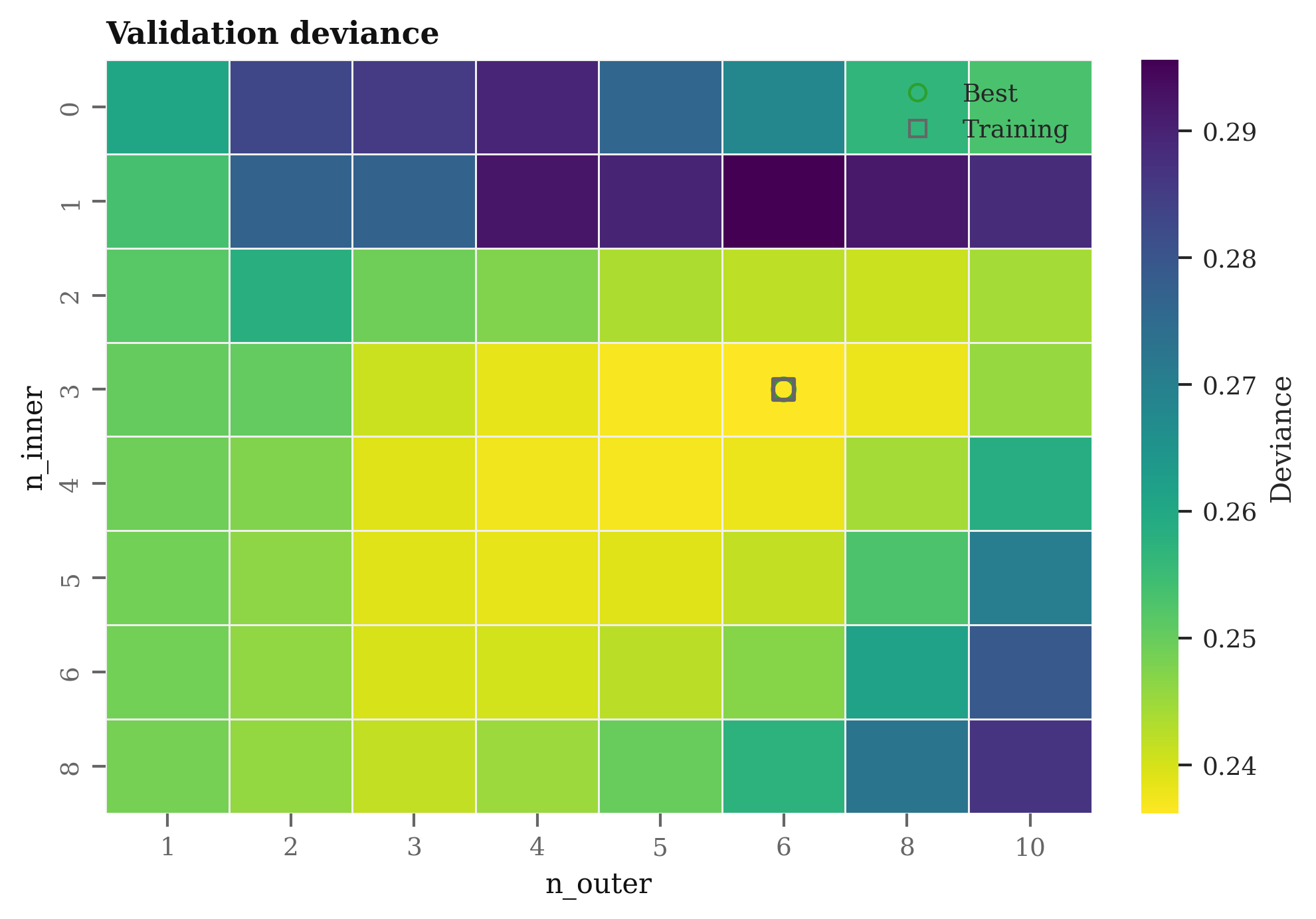}
\caption{Validation deviance heatmap over recursion settings. Outer steps $T\in\{1,2, 3, 4, 5, 6, 8 ,10\}$ and inner iterations $m\in\{0,1,2,3,4,5,6,8\}$ are evaluated on the 10\% validation subset; markers denote the best validation and training configurations, respectively. }
\label{fig:trial56_testtime}
\end{figure}

We evaluate test-time recursion by varying outer steps $T$ and inner iterations $m$ on a held-out validation subset (10\% of the training set, random sample). For each $(T,m)$ we run the recursive layer starting from the validation tokens, compute predictions from the final $\ba$ token and exposure, and report the Poisson deviance loss, see Figure \ref{fig:trial56_testtime}. The validation minimum occurs at $T=6$, $m=3$ with a Poisson deviance loss of $23.890\times 10^{-2}$, close to the Optuna selection score. Applying this configuration on the test set yields a Poisson deviance loss of $23.650\times 10^{-2}$ for the best single model; the nagging ensemble remains at $23.589 \times 10^{-2}$.

\section{Linearized Tab-TRM}
\label{sec:linear_tabtrm}

The interpretability analyses of Section~\ref{sec:trial56_interpretability} revealed that the per-step updates $\Delta\ba^{(t)}$ and $\Delta\bz^{(t)}$ are well-approximated by linear maps in the learned embedding space, with $R^2$ values exceeding $0.8$ across recursion steps. This near-linearity motivates a natural question: \emph{What happens if we enforce exact linearity in the update networks?} In this section we explore a fully linearized variant of Tab-TRM, which provides both a simpler model and a concrete instantiation of the state-space interpretation developed in Section~\ref{subsec:state_space}.

\subsection{Linear Update Architecture}
\label{Linear Update Architecture}
In the linearized model, we remove all nonlinear activations from the update networks $f_z$ and $f_a$, replacing them with single affine maps. The feature encoders (piecewise-linear embeddings for continuous variables, entity embeddings for categorical variables) and the FNN decoder remain unchanged, only the recursive core becomes linear.

Let $\bar{\be} = \frac{1}{L}\sum_{\ell=1}^L \be_\ell$ denote the mean feature token. The inner-loop updates become
\begin{equation}
\bz^{(t,s+1)} = \bz^{(t,s)} + \bW_{zz}\bz^{(t,s)} + \bW_{az}\ba^{(t)} + \bW_f \bar{\be} + \bb_z,
\qquad s=0,\ldots,m-1,
\label{eq:linear_z_update}
\end{equation}
and the outer-loop update is with $\bz^{(t+1)}=\bz^{(t,m)}$
\begin{equation}
\ba^{(t+1)} = \ba^{(t)} + \bW_{aa}\ba^{(t)} + \bW_{za}\bz^{(t+1)} + \bb_a.
\label{eq:linear_y_update}
\end{equation}
These updates are linear in the state $(\ba, \bz)$ and in the feature tokens. With LayerNorm disabled, the recursion is an exact linear state-space model of the form~\eqref{eq:linear_state_space}; with LayerNorm retained, the dynamics are a linear map followed by normalization (piecewise-linear). The observation equation remains the FNN decoder with exponential output activation
\[
\hat{\mu}_i = v_i \, \exp\left(f_o(\ba_i^{(T)})\right) = v_i \, F_\theta(\bx_i, \bc_i),
\]
so the overall model is linear in latent dynamics with a GLM-style nonlinear readout.

\subsection{Predictive Performance}
\label{subsec:linear_results}

To test whether the near-linearity observed in Section~\ref{sec:trial56_interpretability} is sufficient for prediction, we trained the fully linearized Tab-TRM variant (nonlinear activations removed from $f_z$ and $f_a$; encoder and decoder unchanged). Using a 10-run nagging ensemble, the test Poisson deviance losses ranged from $23.695 \times 10^{-2}$ to $23.780 \times 10^{-2}$ (mean $23.742 \times 10^{-2}$), with an ensemble deviance of $23.698 \times 10^{-2}$ and the best single run at $23.695 \times 10^{-2}$. While these results are slightly worse than the full Tab-TRM ensemble studied in  Section~\ref{sec:results}, nonetheless, it is remarkable that a model consisting of multivariate linear models can achieve reasonable performance. This suggests that the recursive learning in Tab-TRM, which performs function composition in latent space, is a powerful mechanism for tabular machine learning.

One way of understanding the success of this linear version of the TRM model is to consider that, in the optimal parameter settings of the TRM model found here, the tokens  $(\ba, \bz)$ and the feature tokens are embedded into a very high-dimensional space ($d=28$). This can be thought of a fine-grained binning of the continuous covariates via the numerical embeddings and, likewise, the categorical covariates are embedded into a very high-dimensonal space. These fine-grained input embeddings adequately capture the complexity of the French MTPL data and the TRM merely needs to select and rearrange this fine-grained representation so that it is suitable for prediction. Exactly for this reason, non-linearity in the TRM recursion is not important, in other words, it is easy to extract the relevant information from these large embeddings. To test this hypothesis, we reran the TRM model with the same hyperparameters but with $d=4$ (i.e., a low-dimensional embedding). The test Poisson deviance losses ranged from $23.644 \times 10^{-2}$ to $23.737 \times 10^{-2}$ (mean $23.706 \times 10^{-2}$), with an ensemble deviance of $23.651 \times 10^{-2}$ and the best single run at $23.644 \times 10^{-2}$. Reestimating the correlation of the updates $\Delta\ba^{(t)}$ and $\Delta\bz^{(t)}$ with the feature tokens, we find that, in this case of lower-dimensional reprsentations, these updates are not nearly as well approximated by linear maps as before, with significant reductions in the $R^2$ values to values of apprximately $0.8$ and $0.5$ for the $\Delta\ba^{(t)}$ and $\Delta\bz^{(t)}$ updates, respectively. We conclude that, while the recursion in the TRM model can deal with complex non-linear updates to extract relevant information from embeddings if needed, nonetheless, in the case of the French data a more optimal approach is to use larger embeddings to model the complexity and rely on linear recursive updates.

\subsection{State-Space Diagnostics}

The linear formulation permits direct analysis of the transition dynamics. Following the derivation in Section~\ref{subsec:state_space}, we can express the system as
\[
\bs^{(t+1)} = \bA\,\bs^{(t)} + \bB\,\bar{\be} + \bc,
\]
where $\bs^{(t)} = [\ba^{(t)}, \bz^{(t)}]^\top$ is the stacked state vector and $\bA$, $\bB$, $\bc$ are the block matrices defined in~\eqref{eq:linear_state_space}.

\begin{figure}[htbp]
\centering
\includegraphics[width=0.5\linewidth]{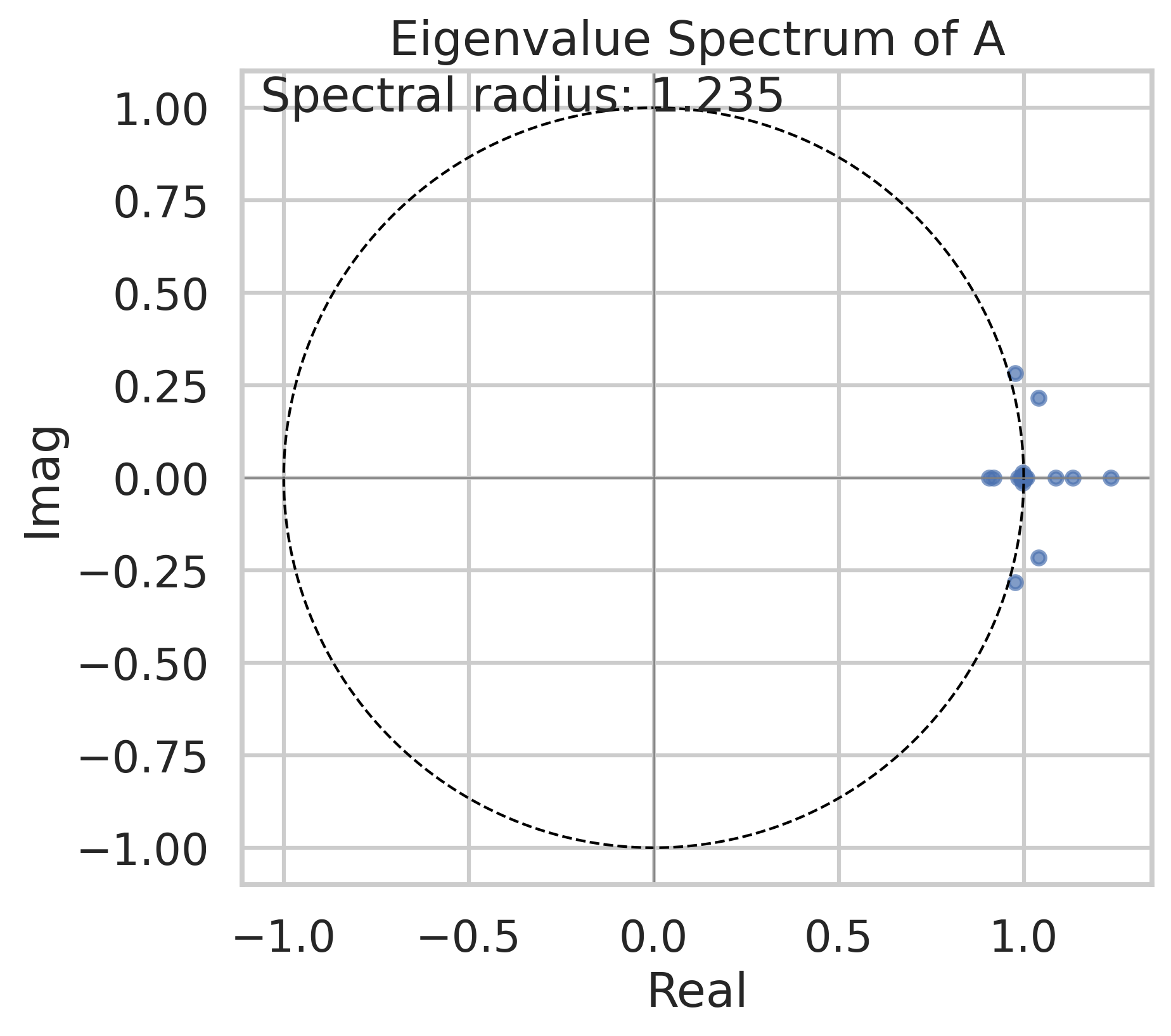}
\caption{Eigenvalue spectrum of the linearized Tab-TRM transition matrix $\bA$. The unit circle is shown for reference. The spectral radius of $1.44$ indicates that the dynamics are not strictly contractive, supporting the interpretation of the recursion as a finite-step refinement rather than convergence to a fixed point.}
\label{fig:linear_eigs}
\end{figure}

\begin{figure}[htbp]
\centering
\includegraphics[width=0.6\linewidth]{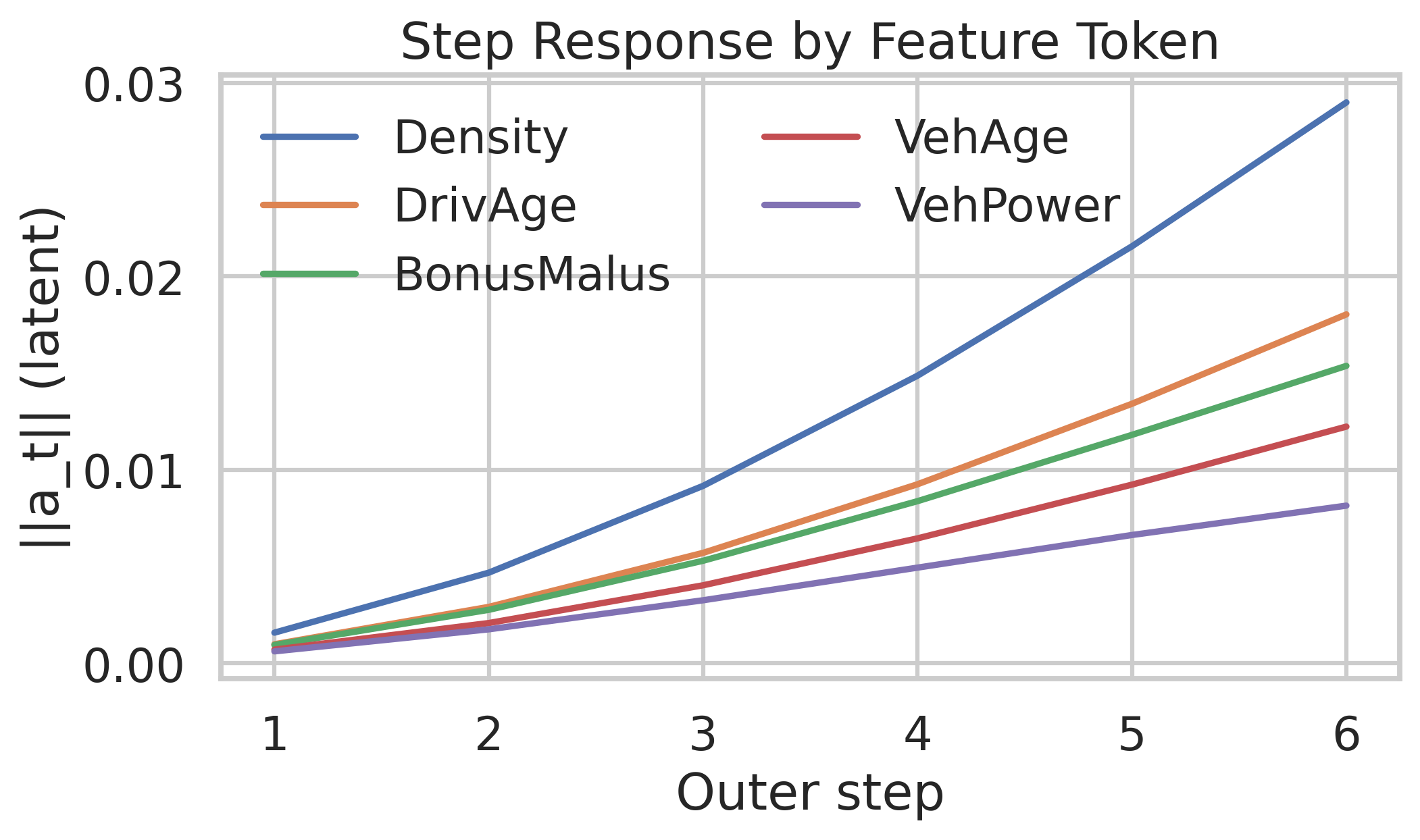}
\caption{Step responses of the linearized Tab-TRM to individual feature tokens. Each curve shows the evolution of the answer token norm $\|\ba^{(t)}\|$ over outer steps when only one feature token is active. Features with larger step responses have stronger immediate influence on the predicted rate.}
\label{fig:linear_step_response}
\end{figure}

\begin{figure}[htbp]
\centering
\includegraphics[width=0.6\linewidth]{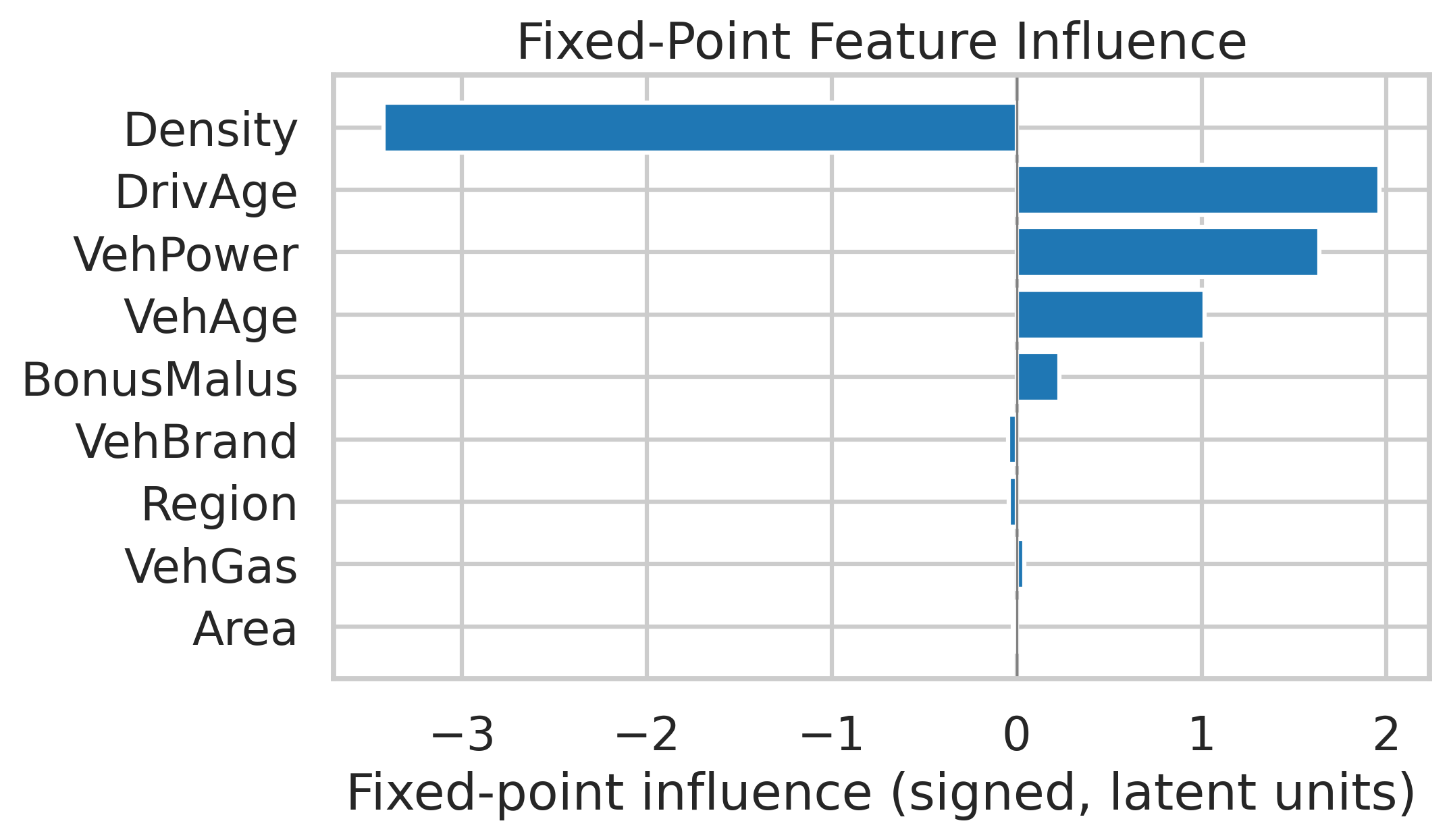}
\caption{Steady-state feature influence for the linearized Tab-TRM, computed via $(\bI - \bA)^{-1}\bB$ and projected onto the mean answer direction. Bars show signed contributions: positive values indicate features that increase the predicted claim frequency, negative values indicate features that decrease it.}
\label{fig:linear_fixed_point}
\end{figure}

Figure~\ref{fig:linear_eigs} shows the eigenvalue spectrum of the effective outer-step transition matrix $\bA$. The spectral radius is $1.44$, implying that the linearized dynamics are not strictly contractive. The recursion therefore acts as a finite-step refinement rather than a convergence-to-fixed-point mechanism.
Figure~\ref{fig:linear_step_response} reports the step response of the linearized model to individual feature tokens, showing how each encoded feature drives the evolution of the answer token $\ba$ over outer steps. Features with larger step responses have a stronger immediate influence on the predicted rate. Figure~\ref{fig:linear_fixed_point} computes the steady-state feature influence via $(\bI - \bA)^{-1}\bB$, which gives the long-run contribution of each feature direction to the latent answer assuming the dynamics were to continue indefinitely. The signed bars reveal which features push the predicted rate up versus down.

\subsection{Implications}

These results reinforce the interpretability narrative developed throughout this paper. Tab-TRM behaves like a linear dynamical system in a learned basis, with the nonlinear feature encoders defining that basis and the recursive core applying repeated linear refinements. The strong performance of the linearized model (within $0.1 \times 10^{-2}$ of the full model) demonstrates that the nonlinear depth of the update networks is not essential for predictive accuracy. Instead, the key inductive bias lies in the recursive state-space structure itself.
This architecture of  nonlinear encoding, linear latent dynamics and nonlinear decoding is reminiscent of classical state-space models in control theory and time-series analysis, but here learned end-to-end from insurance data. When LayerNorm is enabled inside the recursion, the linear diagnostics are approximate but remain informative, since LayerNorm preserves linear directions while normalizing the scale.

\section{Discussion}
\label{sec:discussion}

From a machine-learning perspective, Tab-TRM is closely related to classical recurrent neural network (RNN) architectures, but with a structure that is particularly amenable to actuarial interpretation.
If we unroll the recursive core of Tab-TRM over its $m$ inner iterations and $T$ outer blocks, we obtain a computation graph that resembles a RNN unrolled over $mT$ time steps: a fixed ``cell'' (the tiny feed-forward reasoning core) is applied repeatedly to a hidden state. The key difference is that the hidden state here is split into two parts, $(\ba,\bz)$, with a specific semantic role, namely, $\bz$ is the scratchpad that accumulates reasoning about the features, and $\ba$ is the candidate answer.

Gated architectures such as LSTMs and GRUs were introduced to mitigate vanishing gradients by allowing networks to learn when to remember, forget and update different components of the hidden state. Conceptually, Tab-TRM takes a complementary route: it keeps the state \emph{very small} (just two tokens) and gains depth through recursion and parameter sharing rather than heavy gating. The pair $(\ba,\bz)$ plays a role reminiscent of an LSTM's pair of states (cell state and hidden state), in that one component acts as long-lived memory and the other as a more immediate predictor (naturally through the different updating speeds).

Classical deep learning for tabular data often tries to improve performance by stacking more feed-forward layers. Both HRM and TRM suggest an alternative, namely, keep the network tiny and reuse it many times, using recursion and deep supervision to emulate large effective depth without the parameter blow-up and overfitting risk of very wide or very deep FNNs; see \citet{bai2019deq}. In our setting, this is particularly attractive because insurance datasets, even when they contain hundreds of thousands of policies, may still be small relative to the capacity of modern deep networks. By placing most of the ``depth'' into the recursion rather than into separate parametrized layers, Tab-TRM lives in the same conceptual space as equilibrium and recurrent models, but with a design and loss function that actuaries are already familiar with.

Both the original TRM and our Tab-TRM experiments reveal a \emph{less is more} phenomenon: smaller networks with more recursion often outperform larger networks with fewer recursion steps. This can be explained in several ways. From an overfitting perspective, when training data are limited, large networks quickly memorize the training set, whereas constraining the network to be tiny and reusing its parameters across many recursion steps effectively regularizes the model while still achieving large effective depth. From an architectural perspective, parameter sharing across recursion steps acts as a strong inductive bias, forcing the network to learn a general-purpose ``improvement operator'' rather than separate transformations for each layer. Finally, viewed from an algorithmic standpoint, the recursive structure encodes a prior that the optimal prediction can be reached through iterative refinement.

\section{Conclusion}
\label{sec:conclusion}

We have introduced Tab-TRM (Tabular-Tiny Recursive Model), an architecture that brings the recursive latent reasoning paradigm of TRMs to tabular insurance pricing. Architecturally, we have demonstrated how to tokenize tabular insurance data via categorical embeddings and piecewise-linear continuous encodings and how to augment the resulting token sequence with learned answer ($\ba$) and reasoning ($\bz$) prefix tokens, thereby enabling TRM-style iterative refinement in a classical Poisson GLM setting. Empirically, on a large French MTPL portfolio, Tab-TRM achieves the best test Poisson deviance score with fewer trainable parameters than comparable architectures, confirming that the ``less is more'' principle extends from symbolic reasoning to noisy tabular regression. Conceptually, we have established parallels between Tab-TRM's recursive computation and classical actuarial procedures such as GLMs,  IRLS and modern tabular machine learning in the form of GBMs, making the architecture interpretable to practitioners and highlighting its role as a bridge between traditional actuarial workflows and modern latent recurrent architectures.

We have highlighted the close connection of the Tab-TRM to state-space models. In the case of larger embedding dimensions, the learned state-space dynamics is almost linear, as the nonlinearity can be taken care off by the embeddings and the readout.

Several directions remain for future investigation. The current work focuses on claim frequency; extending Tab-TRM to severity distributions e.g.,  gamma or log-normal, and to combined frequency-severity models is a natural next step. Insurance portfolios have inherent temporal structure: policy years, claims history, and development triangles. Incorporating this structure into the recursive framework, e.g.,  by carrying $(\ba,\bz)$ across time periods, could enable more sophisticated experience rating schemes. Finally, a formal analysis of Tab-TRM's approximation properties and generalization behavior would provide deeper understanding of when and why recursive reasoning helps.

\section*{Acknowledgements}

We kindly thank Darren Cohen from insureAI for the suggestion to consider TRM in the context of insurance pricing.

\bibliographystyle{apalike}

\appendix

\section{Algorithm Pseudo Code}
\label{app:algorithm}

See Algorithm \ref{alg:tabtrm}.

\begin{algorithm}[htbp]
{\footnotesize
\caption{Tab-TRM Forward Pass}
\label{alg:tabtrm}
\begin{algorithmic}[1]
\REQUIRE Policy features $(\bx_i, \bc_i)$, exposure $v_i$
\REQUIRE Embedding functions $\{\phi_j\}_{j=1}^{p_r}$ for continuous and $\{\psi_j\}_{j=1}^{p_c}$ for categorical features
\REQUIRE Initial tokens $\ba^{(0)}, \bz^{(0)}$
\REQUIRE Networks $f_z, f_a, f_o$
\REQUIRE Inner iterations $m$, outer iterations $T$

\STATE \textbf{// Tokenize features}
\FOR{$j = 1$ to $p_r$}
    \STATE Compute $\tilde{x}_{i,j}$ using \eqref{eq:clipped_transform_example}
    \STATE $\be_{i,j} \gets \phi_j(\tilde{x}_{i,j})$ \COMMENT{Continuous embedding}
\ENDFOR
\FOR{$j = 1$ to $p_c$}
    \STATE $\be_{i,p_r+j} \gets \psi_j(c_{i,j})$ \COMMENT{Categorical embedding}
\ENDFOR

\STATE \textbf{// Initialize sequence}
\STATE $\bS_i^{(0)} \gets [\ba^{(0)}, \bz^{(0)}, \be_{i,1}, \ldots, \be_{i,p_r+p_c}]$

\STATE \textbf{// Outer recursion}
\FOR{$t = 1$ to $T$}
    \STATE $\bz_i^{(t,0)} \gets \bz_i^{(t-1)}$,  $\bS_i^{(t,0)} \gets \bS_i^{(t-1)}$

    \STATE \textbf{// Inner recursion (Stage 1: update $\bz$)}
    \FOR{$s = 0$ to $m-1$}
        \STATE $\bu_i \gets \Flatten(\bS_i^{(t,s)})$
        \STATE $\tilde{\bu}_i \gets \LayerNorm(\bu_i)$
        \STATE $\Delta \bz_i^{(t,s+1)} \gets f_z(\tilde{\bu}_i)$
        \STATE $\bz_i^{(t,s+1)} \gets \bz_i^{(t,s)} + \Delta \bz_i^{(t,s+1)}$
        \STATE Update $\bS_i^{(t,s+1)}$ by replacing token 1 with $\bz_i^{(t,s+1)}$
    \ENDFOR

    \STATE $\bz_i^{(t)} \gets \bz_i^{(t,m)}$
    \STATE $\bS_i^{(t,m)} \gets [\ba_i^{(t-1)}, \bz_i^{(t)}, \be_{i,1}, \ldots, \be_{i,p_r+p_c}]$
    \STATE $\tilde{\bS}_i \gets \LayerNorm(\bS_i^{(t,m)})$
    \STATE Extract $\tilde{\ba}_i^{(t-1)}, \tilde{\bz}_i^{(t)}$ as tokens 0 and 1 of $\tilde{\bS}_i$

    \STATE \textbf{// Answer update (Stage 2: update $\ba$)}
    \STATE $\Delta \ba_i^{(t)} \gets f_a([\tilde{\ba}_i^{(t-1)}, \tilde{\bz}_i^{(t)}])$
    \STATE $\ba_i^{(t)} \gets \ba_i^{(t-1)} + \Delta \ba_i^{(t)}$
    \STATE $\bS_i^{(t)} \gets [\ba_i^{(t)}, \bz_i^{(t)}, e_{i,1}, \ldots, e_{i,p_r+p_c}]$
\ENDFOR

\STATE \textbf{// Decode to Poisson log-link prediction}
\STATE $\hat{\lambda}_i \gets \exp(f_o(\ba_i^{(T)}))$
\STATE $\hat{\mu}_i \gets v_i \, \hat{\lambda}_i$

\RETURN $\hat{\mu}_i$
\end{algorithmic}
}
\end{algorithm}

\section{Hyper-parameter Search Space}
\label{app:hyperparams}

See Table \ref{tab:hyperparams}.

\begin{table}[htbp]
\centering
\caption{Hyper-parameter search space for Tab-TRM (Optuna ranges)}
\label{tab:hyperparams}
\begin{tabular}{lcc}
\toprule
\textbf{Hyper-parameter} & \textbf{Range} & \textbf{Type} \\
\midrule
Embedding dimension $d$ & $[16, 60]$ & Integer \\
Inner iterations $n$ & $[1, 8]$ & Integer \\
Outer iterations $T$ & $[1, 6]$ & Integer \\
$f_z$ hidden layers & $[0, 5]$ & Integer \\
$f_a$ hidden layers & $[0, 5]$ & Integer \\
$f_z$ hidden units & $[16, 128]$ & Integer \\
$f_a$ hidden units & $[16, 128]$ & Integer \\
Dropout (core) & $[0.05, 0.5]$ & Float \\
Dropout (decoder) & $[0.05, 0.5]$ & Float \\
$\ell_1$--$\ell_2$ strength & $[0, 10^{-4}]$ & Float \\
Weight decay & $[10^{-3}, 4 \times 10^{-2}]$ & Float \\
Learning rate & $[10^{-4}, 2 \times 10^{-2}]$ & Float \\
Adam $\beta_2$ & $[0.9, 0.99]$ & Float \\
\bottomrule
\end{tabular}
\end{table}

\section{LLM Reasoning Background}
\label{app:llm_background}

This appendix provides additional context on LLM reasoning methods that motivate the latent recursion approach adopted in Tab-TRM.

\paragraph{Chain-of-thought prompting and its limitations.}
Chain-of-thought (CoT) prompting explicitly asks the model to ``think step by step'' and emit a natural-language solution trace before giving the final answer \citep{wei2022cot}. This simple intervention can dramatically improve accuracy on many benchmark datasets, but it comes with costs: long responses, higher latency, sensitivity to the phrasing of the prompt, and the risk that the generated reasoning is partially incorrect even when the final answer happens to be right.

\paragraph{Post-training methods for reasoning.}
On the training side, there is now a spectrum of \emph{post-training} methods that aim to enhance reasoning by shaping the model's intermediate computations. One line of work performs supervised fine-tuning (SFT) on curated or synthetic CoT corpora, treating full reasoning traces as targets rather than only the final answer. A more recent and influential approach is \emph{reinforcement learning with verifiable rewards} (RLVR) \citep{lambert2025rlvr}, which trains models using reinforcement learning where rewards are based on objective, programmatically verifiable criteria, e.g., whether a mathematical answer is correct or whether generated code passes a test suite. The DeepSeek-R1 model \citep{deepseek2025r1} publicly demonstrated this approach, showing that large-scale reinforcement learning with rule-based rewards for accuracy and format can induce emergent CoT reasoning without extensive supervised fine-tuning on reasoning traces.

\paragraph{Reasoning benchmarks.}
A particularly challenging benchmark for evaluating reasoning capabilities is the \emph{Abstraction and Reasoning Corpus} (ARC), introduced by \citet{chollet2019arc} as a measure of general fluid intelligence. ARC tasks present small coloured grids where the solver must infer an abstract transformation rule from a handful of input-output examples and apply it to a new test input. The benchmark is designed to require core human cognitive priors (objectness, goal-directedness, symmetry, counting), rather than pattern-matching on large corpora. Similar grid-based reasoning benchmarks include Sudoku-Extreme (completing $9 \times 9$ puzzles with minimal initial clues) and Maze-Hard (navigating $30 \times 30$ grid mazes). These benchmarks share a common structure: discrete symbolic inputs arranged on a grid, a need for multi-step logical inference, and exact sequence-to-sequence prediction.

\paragraph{Test-time compute.}
To tackle such benchmarks, CoT prompting at inference is often combined with \emph{test-time compute} (TTC): instead of sampling a single CoT, one draws many candidate traces and either takes a majority vote over the final answers or selects the trace with the highest estimated reward \citep{snell2024ttc}. This approach can outperform a single-pass baseline without increasing the number of model parameters, but it further inflates inference cost and still relies on the model's ability to generate coherent natural-language explanations, which may not align with the underlying computation.

\end{document}